\documentclass{vgtc}                          

\graphicspath{{figures/}{pictures/}{images/}{./}} 

\usepackage{amsmath}
\usepackage{times}                     

\onlineid{1020}

\vgtccategory{Research}

\vgtcinsertpkg

\title{Seeing the Many: Exploring Parameter Distributions Conditioned on Features in Surrogates}

\author{Xiaohan Wang\\ %
        \scriptsize Vanderbilt University %
\and Zhimin Li\\ %
     \scriptsize Vanderbilt University %
\and Joshua A. Levine\\
     \scriptsize The University of Arizona %
\and Matthew Berger\thanks{e-mail: matthew.berger@vanderbilt.edu}\\ %
     {\scriptsize Vanderbilt University }}

\abstract{
Recently, neural surrogate models have emerged as a compelling alternative to traditional simulation workflows.
This is accomplished by modeling the underlying function of scientific simulations, removing the need to run expensive simulations.
Beyond just mapping from input parameter to output, surrogates have also been shown useful for inverse problems: output to input parameters.
Inverse problems can be understood as search, where we aim to find parameters whose surrogate outputs contain a specified feature.
Yet finding these parameters can be costly, especially for high-dimensional parameter spaces.
Thus, existing surrogate-based solutions primarily focus on finding a small set of matching parameters, in the process overlooking the broader picture of plausible parameters.
Our work aims to model and visualize the distribution of possible input parameters that produce a given output feature.
To achieve this goal, we aim to address two challenges: (1) the approximation error inherent in the surrogate model and (2) forming the parameter distribution in an interactive manner.
We model error via density estimation, reporting high density only if a given parameter configuration is close to training parameters, measured both over the input and output space.
Our density estimate is used to form a prior belief on parameters, and when combined with a likelihood on features, gives us an efficient way to sample plausible parameter configurations that generate a target output feature.
We demonstrate the usability of our solution through a visualization interface by performing feature-driven parameter analysis over the input parameter space of three simulation datasets. Source code is available at \url{https://github.com/matthewberger/seeing-the-many}}


\graphicspath{{figs/}{figures/}{pictures/}{images/}{./}} 

\usepackage{tabu}                      
\usepackage{booktabs}                  
\usepackage{lipsum}                    
\usepackage{mwe}                       
\usepackage{subcaption}

\usepackage{amssymb}
\usepackage{enumitem}

\usepackage{xcolor}

\begin{document}


\firstsection{Introduction}

\maketitle

Numerical simulations are foundational to the understanding of numerous real-world phenomena.
Domain scientists / engineers often work with simulations by first establishing a governing set of parameters that they wish to study, and upon running the simulation for a given parameter configuration they can then inspect the output, often a set of fields defined over a spatial domain.
Effective visual analysis of a simulation ensemble, where numerous simulations are run under a variety of parameter settings, is a longstanding goal in the visualization community~\cite{wang2018visualization}.


In order to facilitate efficient exploration of simulation ensembles, a more recent trend has been to employ surrogate models~\cite{shi2022vdl,shen2024surroflow,chen2025explorable}.
These often take the form of neural networks~\cite{sitzmann2020implicit} that aim to approximate the computationally expensive mapping from input parameters to simulation outputs.
These surrogates replace repeated full-fidelity simulations with near-instantaneous queries, from parameter input to field output, thus supporting interactive parameter space exploration.


Yet in addition to this forward mapping, e.g. parameter to output, surrogates are also beneficial for tackling inverse problems~\cite{mirghani2012enhanced}.
An inverse problem is defined as determining the input parameters that would produce either a desired simulation output, or some quantity derived from the output, e.g. maximizing lift-to-drag ratio in aerodynamics.
Extensive prior work has applied neural network surrogates to inverse design tasks, including aerodynamic design~\cite{wu2024compositional}, materials discovery~\cite{challapalli2021inverse}, and similar engineering objectives~\cite{mirghani2012enhanced}.
Likewise, the visualization community has also begun exploring using neural surrogates in inverse problem settings~\cite{chen2025explorable,shen2024surroflow}.

Existing approaches to inverse design focus on retrieving a small set of optimal solutions.
However for a given output quantity, specifically a feature identified in a visualized field, a wide range of possible solutions could exist.
This broader set of parameter configurations corresponds to surrogate outputs that give approximate, rather than exact or locally optimal, matches with the output feature.
Displaying this information is valuable in giving a ``bigger picture'' on the parameter space, and supports a feature-driven way of identifying patterns, trends, and relationships among parameter variables -- tasks central to visualizing ensembles~\cite{wang2018visualization}.
To this end, in this paper we take on a \emph{distributional} perspective.
Instead of retrieving a handful of input parameters, our goal is to both model, and visualize, the distribution of possible input parameters that could lead to a given output.



This distributional approach, however, introduces two critical challenges.
First, surrogate models serve as approximations to simulation runs and thus inevitably contain error.
The resulting distribution should be aware of these imperfections, where we only wish to consider parameters where the surrogate gives accurate predictions.
Secondly, forming the distribution must be fast enough to support interactive visual analysis.
We address these challenges in the context of Bayesian inference, making two key contributions:
\begin{enumerate}[leftmargin=.4cm]
    \item We propose a method for density estimation in neural field surrogates that can be used to model error. In relation to standard uncertainty estimates~\cite{lakshminarayanan2017simple}, our method is simple, faster to compute, and comparable in being predictive of surrogate error.
    \item The density estimate represents our prior belief on parameters, which when combined with a likelihood on a user-supplied feature, allows us to formulate the inverse problem as that of drawing samples from a posterior distribution. By using Hamiltonian Monte Carlo for sampling, we show that obtaining a good approximation to the posterior can be done fast enough, and in a progressive manner, to minimize latency.
\end{enumerate}
We demonstrate our method's advantages with an interface design that visualizes parameter distributions.
We illustrate how our design allows one to make numerous findings about nontrivial simulation parameter spaces in an interactive, feature-driven manner.


\section{Related work}

We focus our discussion of related work on the use of surrogates for modeling ensembles in visualization and machine learning, so as to highlight the different applications and settings they are used.  Finally, we briefly discuss visual interface design for ensembles, with a particular focus on exploring parameters.

\subsection{Surrogate Models in Visualization}

The use of surrogate models in visualization often focuses on mitigating computational and storage limitations. 
In this work we use surrogates that encode the mapping between simulation parameters and simulation outputs such that the mapping generalizes from discrete simulation data (sampled at specific combinations of parameters and positions) to field outputs.
However, surrogates can model many other modules within visualization pipelines.
For example, work such as Berger et al.~\cite{berger2018generative}, He et al.'s InSituNet~\cite{he2019insitunet}, and Shi et al.'s VDL-Surrogate~\cite{shi2022vdl} explore the use of surrogates of the visualization process, mapping output images (rather than data) from  visualization parameters (such as transfer function and camera, as in Berger et al.), combinations of visualization and simulation parameters (He et al.), or capturing view-dependent information along rays (Shi et al.)
These works allow for rapid exploration of visualization outputs at different configurations, without the need to query the original volumetric data.  
Generally such models can be viewed as forward processes, although there is limited support for guiding the use towards interesting regions of parameter space.
Furthermore, asking for a different aspect of the data not modeled by the surrogate might require building a new surrogate from scratch.

Early work in the visualization community constructed custom data surrogates for a variety of use cases, often employing representations based on deep learning.
For example, Hazarika et al.'s NNVA~\cite{hazarika2019nnva} constructs a surrogate model, used to map simulation parameters to the entire output (400 values), as the backend data representation for a visual analysis system.
Predicting values at specific locations discards the adjacency and relationships between individual values, so instead of predicting discrete values, Shi et al.'s GNN-Surrogate~\cite{shi2022gnn} uses graph neural networks more suitable for unstructured mesh representations.
In either example, predicting an entire dataset for a set of simulation parameters is costly, as the size of the network must be scaled with the size of output representation.
Instead, others have explored the use of coordinate-based MLP architectures~\cite{sitzmann2020implicit,tancik2020fourier}, originally introduced to the visualization community as a vehicle for data compression~\cite{lu2021compressive}.
Such \emph{implicit neural representations}, or INRs, predict output values for individual spatial position rather than predicting the entire field.
Work such as Han and Wang's CoordNet~\cite{han2022coordnet} demonstrates the potential of INRs in a variety of visualization tasks.
We also elect to use an INR as a surrogate representation, factoring in both spatial position and parameters as the input to our model.

\subsection{Using Surrogates for Inverse Problems}

When a surrogate models the forward process, the setup generally optimizes for rapid evaluation of the output for a given set of input parameters.  
As a result, often the underlying surrogate is ill-suited for parameter exploration since the inverse task (going from output back to parameters) is not easily achieved.  
A full inverse design pipeline using surrogates requires different machinery.
Recently, neural surrogates have been studied for inverse design tasks in multiple domains~\cite{challapalli2021inverse,mirghani2012enhanced}. 
We highlight the method of Wu et al.~\cite{wu2024compositional} which frames the inverse design problem as an optimization process to find the best parameters that, when evaluated on the surrogate, achieve a target design.
By replacing the surrogate with a diffusion model (specifically, a generative energy function), they introduce a compositional perspective that aims to capture whether the produced parameters are within the input distribution.  
While we utilize different tools in our work, we share a similar perspective in that our Bayesian framework narrows the search space to parameters within the distribution.

Although numerous machine learning methods study the inverse problem, the visualization literature on inverse surrogates has limited recent work.
Surroflow~\cite{shen2024surroflow} is a normalizing flow-based surrogate model that enables invertible prediction between simulation output and input for parameter exploration. 
While normalizing flow offers a mechanism for invertibility, it frames the inverse problem as defining a constraint on the entire field.
Instead, our approach is more flexible, as it enables parameters to be retrieved based on a selected portion of the simulation (e.g., a slice of a field feature).
Explorable INR~\cite{chen2025explorable} uses an INR-based surrogate, similar to ours, but frames the inverse problem through the lens of uncertainty propagation.
While a feature can be specified in a local region, gradient descent is used to infer matching parameters, which limits the method to producing candidate matches, instead of capturing a complete description of the set of parameters that match a target feature.

\begin{figure*}
    \centering
    \includegraphics[width=\linewidth]{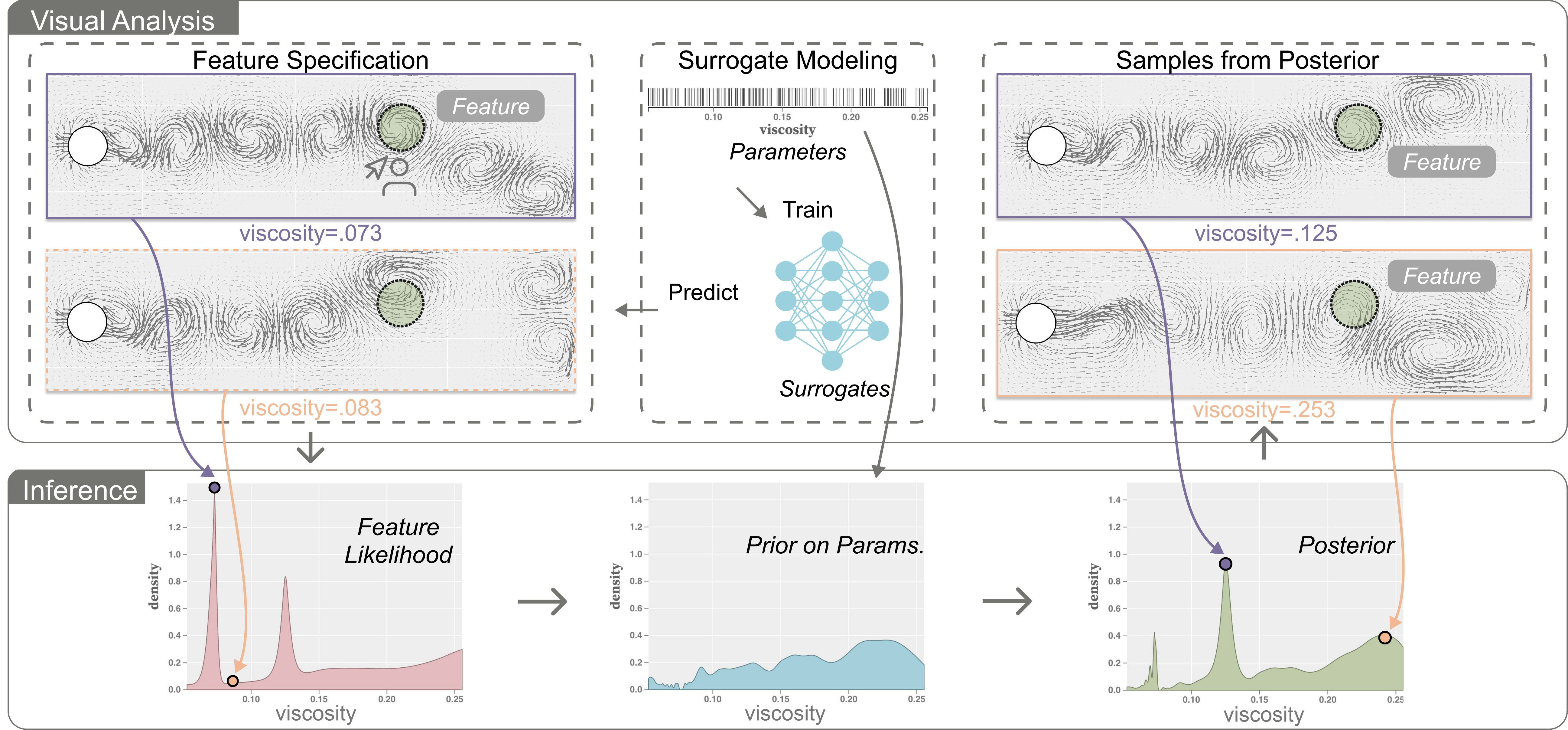}
    \caption{We illustrate our context use: the user wishes to infer viscosity from observed vortices. We sample the parameter space, train a neural surrogate, and obtain a prior over viscosity. We use surrogates to visualize flow fields, where the user marks a target vortex feature. We compute the likelihood of that feature for each viscosity and combine it with the prior to form a posterior, from which we draw samples. On the right, the upper field has higher posterior weight and better matches the user-selected feature. 
    For clarity, we show a 1-D parameter, but our method generalizes to higher-dimensional spaces.}
    \label{fig:appr-overview}
\end{figure*}

\subsection{Uncertainty and Surrogates}

Due to both the way ensembles are analyzed and the approximate nature of surrogates, it is natural that downstream visualization processes consider uncertainty~\cite{molnar2024uncertainty}.   
When considering uncertainty coming from a neural surrogate's approximation, standard ML methods such as Monte Carlo dropout~\cite{gal2016dropout} and neural network ensembles~\cite{lakshminarayanan2017simple} are commonly applied, but they also introduce additional training or inference overhead. 
Saklani et al.~\cite{saklani2024uncertainty} show how to apply these measures to convey uncertainty in INRs.
Likewise, Xiong et al.'s RMDSRN~\cite{xiong2024regularized} thoroughly evaluate the use of these uncertainty measures in a scene representation network as it impacts volume rendering. 
Although we recognize the errors introduced by using a surrogate, our goal is not to quantify the error it introduces. 
Instead, we use density estimation to hone in on parameter combinations that give accurate predictions.
An alternative might be to explicitly model the correlation between parameters, as done by Farokhmanesh et al.'s neural dependence fields~\cite{farokhmanesh2023neural}.  

More broadly in the space of machine learning, we take motivation from recent developments in modeling uncertainty in neural radiance fields, or NeRF~\cite{mildenhall2021nerf}.
The interest in uncertainty for NeRF-like models is typically for understanding reconstruction error and guiding view selection.  
For example, Goli et al.'s Bayes’ Rays~\cite{goli2024bayes} utilizes the Laplace approximation to simulate perturbations that generate approximate spatial uncertainty for NeRF.
Likewise, Jiang et al.'s FisherRF~\cite{jiang2023fisherrf} uses Fisher information as a metric to quantify the uncertainty for view selection.
To draw a parallel with our approach: we also use a Bayesian framework and take advantage of Fisher information (specifically, in computing our prior).
In this sense, one can view the camera parameters in NeRF as analogous to the governing simulation parameters in our ensemble (noting that NeRF is not a true ensemble surrogate).
However, we are ultimately interested in capturing uncertainty measures for the inverse problem.
Thus, these works are inspirational for our method, but we consider a different aspect of uncertainty quantification.

\subsection{Ensemble Parameter Exploration and Visualization}

Many in the visualization community have considered visualization for ensembles, regardless of the use of a surrogate.  
In this section, we briefly focus on inverse design aspects around visual parameter exploration, noting that a complete coverage of ensemble visualization is covered in a recent survey by Wang et al.~\cite{wang2018visualization}.
Sedlmair et al.~\cite{sedlmair2014visual} propose a visual parameter space analysis framework encompassing data generation, presentation, and user analysis. 
For parameter space analysis, one conceptual approach is to build groups (clusters) of ensemble members and parameter settings where a visualization allows a user to interact with the grouping~\cite{bruckner2010result,obermaier2015visual,wang2016multi}. 
Our Bayesian framework replaces the notion of groups with a density estimation across parameter combinations.
Others consider how best to visually manipulate a simulation output to achieve inverse design, such as Design by Dragging~\cite{coffey2013design} and Drag and Track~\cite{orban2018drag}.
However, in order to perform an exploration and design with simulations, these approaches can face computational challenges and may limit the process to only working with discrete simulation outputs.
That said, their methodology is of interest in our work as it frames both how we could define a feature and simultaneously interact with the parameter space.

Other works in visualization for ensembles have considered Bayesian perspectives for parameter exploration of ensembles.  
Gosink et al.~\cite{gosink2013characterizing} utilize Bayesian model averaging to extrapolate prediction uncertainty from ground truth weather observations.
Biswas et al.~\cite{biswas2016visualization} estimate sensitivity across different spatial resolutions using Bayesian tools.
In principle, we share common elements with both of these works, but our approach differs significantly in that we focus on a different goal of the inverse problem and produce estimates of density that have different interpretations.

\section{Methods}

\subsection{Problem setup}
\label{sec:setup}

To discuss the problem we aim to solve we first introduce some mathematical preliminaries.
We represent a simulation via (1) a $d$-dimensional parameter space $\mathcal{Z} \subset \mathbb{R}^d$ corresponding to all potential parameter inputs, (2) the domain of a simulation output $\mathcal{X} \subset \mathbb{R}^m$ where dimensionality $m$ captures the spatial and temporal dimensions of the domain, and (3) a function that maps from input parameter to simulation output $f : \mathcal{X} \times \mathcal{Z} \rightarrow \mathbb{R}^n$.
In our setting we assume a simulation output gives a 2D time-varying vector field, thus $m = 3$ for the 2 spatial dimensions \& time, and $n = 2$ to give an output $2$-vector.
The parameter space dimensionality $d$ varies according to the details of the simulation.

We further assume a user-provided \emph{feature}. 
For a vector field, a feature can be as simple as the simulation output at a specified location, to quantities derived from spatial(-temporal) derivatives, or integral curves -- we only stipulate that the feature is differentiable with respect to the input parameter $\mathbf{z} \in \mathcal{Z}$.
To limit the scope of our work, we define a feature from a user-specified spatial neighborhood $X \subset \mathcal{X}$ at a fixed slice of time, where we gather the simulation output at a set of spatial positions within $X$, yielding a collection of 2D vectors.
The size of the neighborhood determines the granularity of the feature, namely, as the neighborhood shrinks to a single point, we obtain just a single vector.
Larger neighborhoods allow us to capture more complex behavior, e.g., regions of laminar flow or vortices within a vector field.


We can now mathematically formulate the aim of this work as finding the following set of parameters:
\begin{equation}
\{\mathbf{z} \in \mathcal{Z} \, | \, d_X(\mathbf{z}, \mathbf{\hat{z}}) \le C\}.
\label{eq:exact}
\end{equation}
The function $d_X$ measures how close simulation outputs in a user-specified parameter configuration $\mathbf{\hat{z}}$ matches a given $\mathbf{z} \in \mathcal{Z}$, over spatial positions sampled within $X$:
\begin{equation}
d_X(\mathbf{z}, \mathbf{\hat{z}}) = \frac{1}{K} \sum_{k=1}^K \lVert f(\mathbf{x}_k, \mathbf{z}) - f(\mathbf{x}_k, \mathbf{\hat{z}}) \rVert,
\label{eq:neighbdist}
\end{equation}
for $K$ number of positions $\mathbf{x}_k$ sampled uniformly at random within $X$.
The parameter $C$ in Eq.~\ref{eq:exact} is the threshold at which we declare a feature match between parameter configurations $\mathbf{\hat{z}}$ and $\mathbf{z}$.
Note that when $C = 0$, we are asking for the features to be equal.
Stated this way, computing $d_X$ requires access to the simulation mapping $f$ for arbitrary input parameters.
In practice, however, typical ensemble simulations cover a finite number of parameter configurations, and thus the mapping $f$ gives rise to a finite number of fields.
Instead, we employ a surrogate model to approximate $f$, which we denote as $f_{\theta}$ with model parameters $\theta$.
We assume the surrogate can be evaluated at any input in the domain and input in parameter space, and we place an additional constraint that the surrogate is differentiable with respect to parameters.
This can take the form of scattered data interpolation~\cite{coffey2013design}, or a coordinate-based neural network~\cite{sitzmann2020implicit, muller2022instant, chen2025explorable}.

Given the surrogate, one may proceed by replacing $f$ with $f_{\theta}$ in Eq.~\ref{eq:neighbdist}.
However, constructing the set (c.f.~Eq.~\ref{eq:exact}) with $f_{\theta}$ presents several difficulties.
First, we must deal with imperfections in the surrogate model.
For certain parameter configurations $\mathbf{z} \in \mathcal{Z}$ the corresponding surrogate model output might give a poor approximation to the ground truth, e.g.~the results of running the simulation for $\mathbf{z}$.
In forming the set, we wish only to retain parameter configurations that are likely to give accurate approximations so as to not mislead the user.
Secondly, the computational expense of forming this set should be minimized.
We aim to support users in interactively specifying features over the domain of the field, and subsequently, receiving the parameter configuration set with minimal latency.
For low-dimensional parameter spaces, e.g.~$d \le 2$, brute-force search is feasible in finding parameters.
Otherwise, naive search becomes computationally prohibitive to ensure interactivity.

In our approach, we address these challenges by adopting tools from Bayesian inference, please see Fig.~\ref{fig:appr-overview} for an illustration.
Specifically, we first define a \emph{feature likelihood}, denoted $p(X | \mathbf{z})$ for some user-defined region $X$.
This represents a score of how likely a parameter configuration matches a user-defined feature.
Secondly, we aim to construct a \emph{prior} on parameters, denoted $p(\mathbf{z})$.
This quantity represents our prior belief on parameters, namely the likelihood that the surrogate model gives an accurate approximation when presented with an input parameter.
We then aim to display the joint probability $p(X, \mathbf{z}) = p(X | \mathbf{z}) \cdot p(\mathbf{z})$ over the parameter space $\mathcal{Z}$.
Regions of the parameter space for which the joint probability is high reflect parameter configurations that are likely to both match a feature \emph{and} give accurate surrogate outputs.
Evaluating this quantity over the entire parameter space remains computationally expensive.
We address this challenge by instead displaying the density of samples drawn from the posterior:
\begin{equation}
    p(\mathbf{z} | X) = \frac{p(X, \mathbf{z})}{p(X)}.
\label{eq:posterior}
\end{equation}
By framing this problem via sampling, a central consideration is how many samples are necessary to give a good approximation to the posterior $p(\mathbf{z} | X)$.
Approximate inference schemes, namely Hamiltonian Monte Carlo~\cite{neal2011mcmc}, give us an efficient means of sampling, assuming the posterior density is concentrated in a small portion of the parameter space.
In subsequent subsections we discuss in more detail the use of neural fields for surrogate modeling (Sec.~\ref{sec:nf}), the specification of features using neural fields (Sec.~\ref{sec:feat_likelihood}), our prior on parameters (Sec.~\ref{sec:prior}), inference computations (Sec.~\ref{sec:inference}), and our visualization design for displaying the results of inference (Sec.~\ref{sec:design}).

\subsection{Neural field surrogate}
\label{sec:nf}

We propose to use coordinate-based neural networks to represent the space of fields produced by a simulation.
Specifically, we utilize a SIREN network~\cite{sitzmann2020implicit} to represent $f_{\theta}$, presenting as input to the model both the coordinates of the domain (e.g.~space and time) and simulation parameters.
The model outputs an element in the range of the field, e.g.~for a 2D time-varying vector field this would be a $2$-vector.
We train the model on a set of simulation runs spanning the parameter space of the simulation, where optimization aims to minimize a mean-squared error loss.
We note that for more efficient inference, other neural field representations could be used~\cite{dupont2022data,bauer2023spatial}, however, to limit the scope of our work, we only consider SIRENs for surrogates.
In the supplemental material, we provide additional details regarding the model specification.

\subsection{Feature likelihood}
\label{sec:feat_likelihood}

We can use our neural field surrogate to define the neighborhood-based feature with Eq.~\ref{eq:neighbdist}.
Instead of using the set computation in Eq.~\ref{eq:exact}, we calculate a feature likelihood $p(X | \mathbf{z})$ as follows:
\begin{equation}
    p(X | \mathbf{z}) \propto \exp\left(-\frac{d_X(\mathbf{z}, \mathbf{\hat{z}})}{C}\right),
    \label{eq:spec}
\end{equation}
where $d_X$ utilizes the neural field for predicting simulation outputs, and the positive scalar $C$ is in correspondence with the tolerance parameter in Eq.~\ref{eq:exact}.
Note that $p(X | \mathbf{z})$ is only proportional to the quantity on the right-hand side, up to a multiplicative constant.

\subsection{Prior on parameters}
\label{sec:prior}

Our prior on parameters $p(\mathbf{z})$ aims to measure how likely the surrogate model produces an accurate output for a given parameter configuration $\mathbf{z} \in \mathcal{Z}$.
Moreover, we aim for the prior computation to be no more expensive than evaluating the feature likelihood.
We define the prior with two criteria.
First, a given parameter configuration should be close to the parameters used in training the neural field surrogate.
Second, the surrogate model output for a parameter configuration, e.g. a field sampled to a set of points in the domain, should be similar to the surrogate model output restricted to the training parameter configurations.
We report high prior belief for a parameter configuration if it remains close to training data parameters, as measured by the input and output of the surrogate model.
We realize these criteria through kernel density estimation.

In detail, we can make concrete the notion of input proximity, or how ``close'' a given input $\mathbf{z} \in \mathcal{Z}$ is with respect to the training data, via the following:
\begin{equation}
	k(\mathbf{z} ; \sigma_s) = \sum_{i = 1}^N \exp\left( - \frac{\lVert \mathbf{z} - \mathbf{z}_i \rVert^2}{\sigma_s^2}\right),
\label{eq:kde}
\end{equation}
where $\mathbf{z}_i , i \in [1,N]$ represents training data parameters, $N$ in total, and we use a Gaussian kernel in the above with bandwidth $\sigma_s$.
Intuitively, if $\mathbf{z}$ is close to other parameters observed during training, then the KDE reports a high density.

Although conceptually straightforward and inexpensive to compute, a limitation with such a KDE is that it discards any knowledge about the surrogate model.
We would thus like to capture a notion of proximity on the surrogate output, not just the input.
One way to measure this is to evaluate the surrogate model to a field of fixed resolution for a presented pair of parameter configurations, and then compute the distance between the resulting fields.
We can more formally express this as follows:
\begin{equation}
	D^2(\mathbf{z}, \mathbf{z}_i) = \sum_{j = 1}^M \lVert f_{\theta}(\mathbf{x}_j, \mathbf{z}) - f_{\theta}(\mathbf{x}_j, \mathbf{z}_i) \rVert^2,
\end{equation}
where the sum is over $M$ elements of the field's domain, namely each point $\mathbf{x}_j$ is sampled uniformly at random from $\mathcal{X}$.
For regions of the parameter space where the mapping $f_{\theta}$ is rapidly changing with respect to input parameters, the reported distance is more likely to be large.
A problem with this approach is that it is computationally prohibitive.
Instead, we use a first-order approximation of $f_{\theta}(\mathbf{x}_j, \mathbf{z})$ to approximate $D^2(\mathbf{z}, \mathbf{z}_i)$:
\begin{equation}
	 \approx D^2_{FIM}(\mathbf{z}, \mathbf{z}_i) = (\mathbf{z} - \mathbf{z}_i)^T \mathbf{J}_{\theta}(\mathbf{z_i})^T \mathbf{J}_{\theta}(\mathbf{z_i}) (\mathbf{z} - \mathbf{z}_i),
\end{equation}
where $\mathbf{J}_{\theta}(\mathbf{z_i}) \in \mathbb{R}^{M \cdot n \times d}$ is the Jacobian matrix corresponding to the surrogate model $f_{\theta}$, with derivatives computed with respect to the parameter dimensions.
The matrix $\mathbf{F}(\mathbf{z}_i) = \mathbf{J}_{\theta}(\mathbf{z_i})^T \mathbf{J}_{\theta}(\mathbf{z_i}) \in \mathbb{R}^{d \times d}$ is the Fisher information matrix (FIM), a positive semi-definite matrix that captures the sensitivity of the surrogate model under small perturbations in the parameter space.
The distance $D^2_{FIM}$ is a squared Mahalanobis distance, and serves as a cheap approximation to the original distance $D^2$, presuming we have access to $\mathbf{F}(\mathbf{z}_i)$.
We can utilize this new distance as part of our KDE in Eq.~\ref{eq:kde}, where we combine distances in the density estimate:
\begin{equation}
	p(\mathbf{z} ; \sigma_s, \sigma_f) \propto \sum_{i = 1}^N \exp\left( - \frac{\lVert \mathbf{z} - \mathbf{z}_i \rVert^2}{\sigma_s^2} - \frac{D^2_{FIM}(\mathbf{z}, \mathbf{z}_i)}{\sigma_f^2} \right).
\label{eq:prior}
\end{equation}
Each matrix $\mathbf{F}(\mathbf{z}_i)$ can be computed and stored in a preprocessing step, making evaluating Eq.~\ref{eq:prior} computationally efficient.

\begin{figure}[!t]
    \centering
    \includegraphics[width=0.9\linewidth]{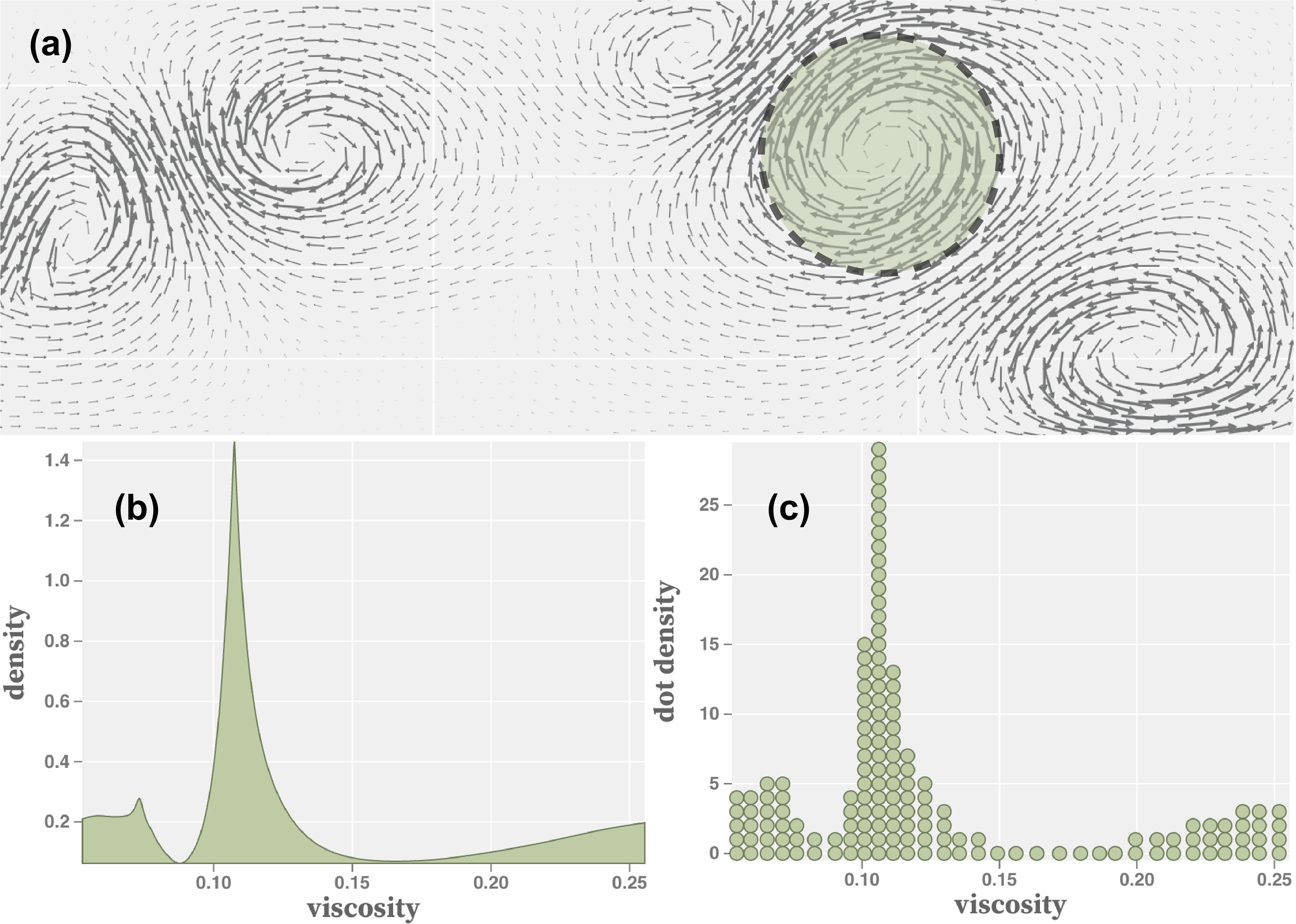}
        \caption{
        Given a user-specified feature of a simulation's output (a), we aim to display the posterior, combining feature likelihood and our prior belief on parameters. 
        Dense evaluation over all parameters (b) scales poorly to higher-dimensional parameter spaces. Instead, we use Bayesian inference to efficiently sample from the posterior (c), wherein sample density approximates the posterior.}
    \label{fig:basic_idea}
\end{figure}

The choice to retain the distance between parameters (c.f.~Eq.~\ref{eq:kde}) is due to the assumptions inherent in $D^2_{FIM}$, namely, the closer the parameter inputs, the better the approximation.
In practice, we set the bandwidth for the FIM distance $\sigma_f$ to be the average of nearest neighbor FIM distances computed over training parameters, while $\sigma_s$ is set to 6 times the average of nearest neighbor (Euclidean) distances.
Thus, when parameters are close by, the FIM distance takes precedence, which is precisely the scenario where we expect $D^2_{FIM}$ to give a good approximation.


\subsection{Approximate inference}
\label{sec:inference}

Given our likelihood and prior, we next consider how to draw samples from the resulting posterior (c.f.~Eq.~\ref{eq:posterior}).
We use Hamiltonian Monte Carlo~\cite{gelman1995bayesian, neal2011mcmc} (HMC) for this purpose, a form of Markov chain Monte Carlo for which gradients of the posterior guide the exploration of the parameter space.
We use the standard leapfrog integration scheme~\cite{neal2011mcmc} to numerically integrate Hamilton's equations.
HMC requires a number of parameters to be set to ensure both convergence to the target posterior and sufficient coverage of it: step size, number of leapfrog steps, and the number of samples collected along a chain.
We tune the step size such that the Metropolis-Hastings (MH) acceptance probability is roughly 50\%, thus making sure that we are sufficiently exploring the density, while not taking steps so large as to violate the MH criterion.
The step size can be set independently of both feature specification and the simulation.
In setting the number of leapfrog steps to 10, we find that, on average, the length of the integrated path is roughly 10\% of the diameter of the parameter space.
This ensures the path sufficiently explores the parameter space while avoiding cycles.
Last, chains of size 100 - 200 steps, with burn-in of 50 steps, give good coverage of the density when multiple chains are run in parallel; we further evaluate these choices in the supplemental material.


The density of the collected samples is intended to be proportional to the posterior density.
We highlight this for a simple example of a 1-dimensional parameter space in Fig.~\ref{fig:basic_idea}, for a fluid flow simulation with an embedded circle boundary.
We fix the circle shape/location and initial conditions, and set the fluid's viscosity as the single parameter that varies over the simulation.
A given specified feature (Fig.~\ref{fig:basic_idea} (a)) is shown as the target neighborhood of vectors that we utilize in forming the feature likelihood, alongside the prior.
Due to the low-dimensional parameter space, it is feasible to compute the posterior density over a dense set of viscosity values,
displayed as an area mark on the left.
The results of HMC are shown on the right, where we display the set of collected samples as a quantile dot plot~\cite{fernandes2018uncertainty}.
Note that the density of samples (number of dots at a given viscosity value) captures the same information that we obtain in the posterior density.
For higher-dimensional parameter spaces this equivalence is essential, since depending on the shape of the posterior density, we can efficiently obtain an accurate approximation to the posterior density via sampling.

The collected set of samples can be used to derive additional information about the simulation's output space.
We compute the standard deviation of the surrogate model's output over the set of parameter configurations collected via HMC.
This is done at each location on a regular grid defined over the spatial domain.
Intuitively, a high standard deviation conveys the regions in space that need to change in order for the specified feature to remain the same.
This represents a generalization of variability visualization methods that consider all simulations of an ensemble~\cite{potter2009ensemble}, where we limit simulations to those which are likely to contain a specified feature.

\begin{figure}
    \centering
    \includegraphics[width=1.0\linewidth]{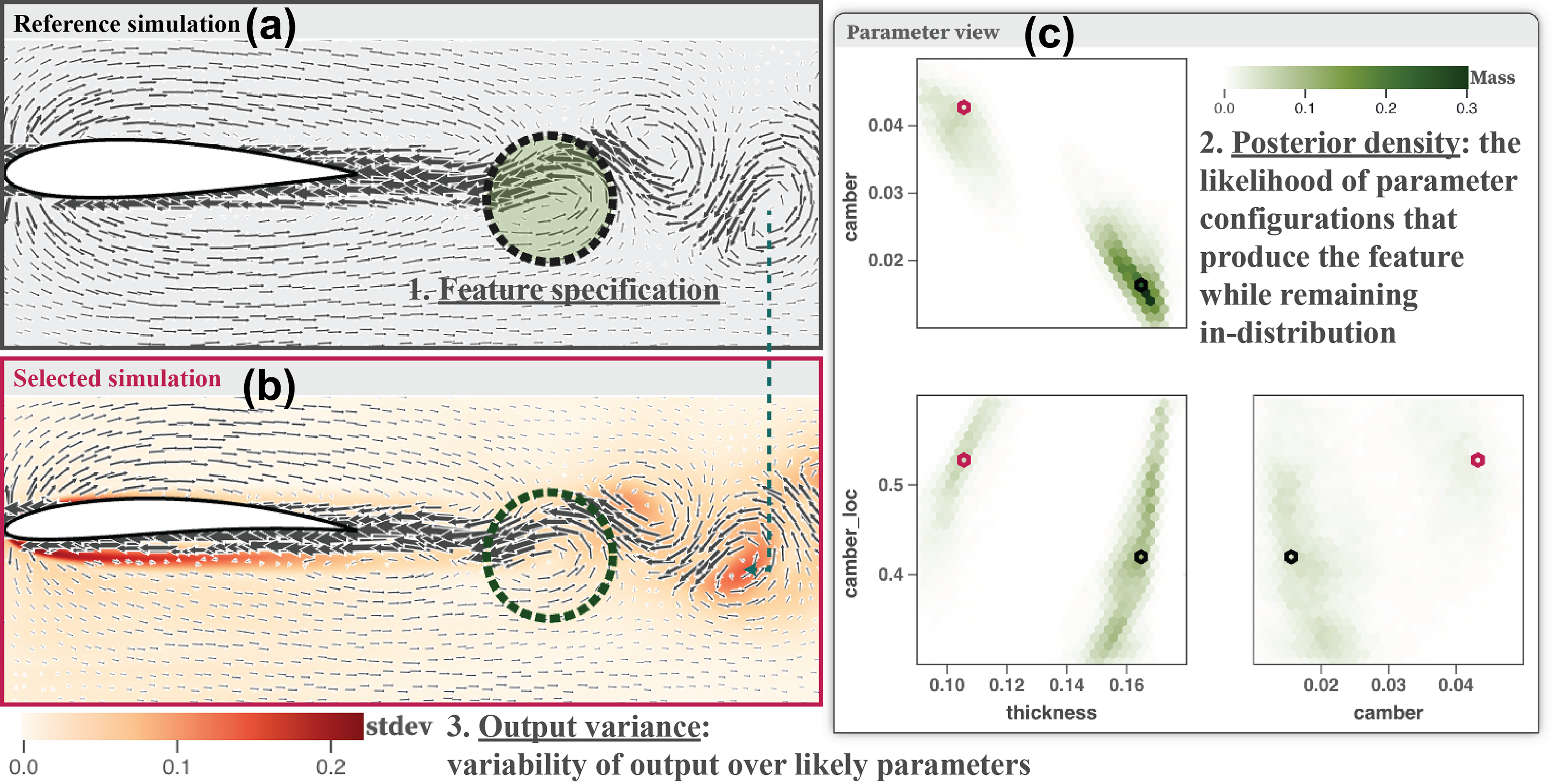}
    \caption{
    Our visualization design for feature-driven exploration of simulations. Users selects a feature via a spatial region of a reference field output by the surrogate model (a). Samples collected via HMC are then shown in a matrix-based view of binned heatmaps encoding sample density over pairs of parameter variables (c). We show output variance (b) across the domain, highlighting where changes are needed to preserve the feature.}
    \label{fig:design}
\end{figure}

\section{Visualization design}
\label{sec:design}

Our visualization design consists of two main views: a view that depicts the surrogate-predicted field for selected parameters, and a view that shows the results of approximate inference, namely the density of the collected samples.
Please see Fig.~\ref{fig:design} for an illustration.
To limit the scope of our work, we limit our field-based visualization to that of vector fields and show a given field via a set of sampled vector glyphs.
We show two types of fields -- one allows for the user to specify neighborhood-based features (Fig.~\ref{fig:design}-a), and the other allows for the display of fields for browsed parameter configurations (Fig.~\ref{fig:design}-b).
By choosing two distinct parameter configurations both assigned as high density, one can then inspect their corresponding fields to observe how the features are preserved, while also viewing the full domain to identify differences in the fields.

The other main view in our design shows a scatterplot matrix-like display of the set of collected samples (Fig.~\ref{fig:design}-c).
Specifically, each subview in the matrix corresponds to a unique pair of parameter variables, where we display the density of the samples via a binned heatmap.
This matrix-based design can be understood as depicting all 2-dimensional marginal distributions of the posterior.

Our interface supports a number of interactions for both specifying features and exploring the sampled parameter configurations.
In the reference view, the user drags a region widget, namely a circle, to specify the neighborhood region on which the feature (c.f.~Eq.~\ref{eq:spec}) is derived.
In theory, the circular widget may span the entire domain, in which case the specified feature corresponds to the full field.
After dragging, we run HMC to collect samples drawn from the posterior.
Once a predefined number of burn-in iterations has passed, we stream in samples collected along each chain, thus giving a progressive visualization of the density.
Upon completion, users may interactively hover over any subview in the matrix layout to update the selected field view (Fig.~\ref{fig:design}-b), while clicking on a given point in a subview will update the field view that supports feature specification (Fig.~\ref{fig:design}-a).
In other words, the parameters are adjustable directly via the interface.
Moreover, within the selected field view we provide a color-based encoding of the output variance, as measured over the collected parameter configurations, shown in the background of the vector glyph view.

\begin{figure}[!t]
    \centering
    \includegraphics[width=1.0\linewidth]{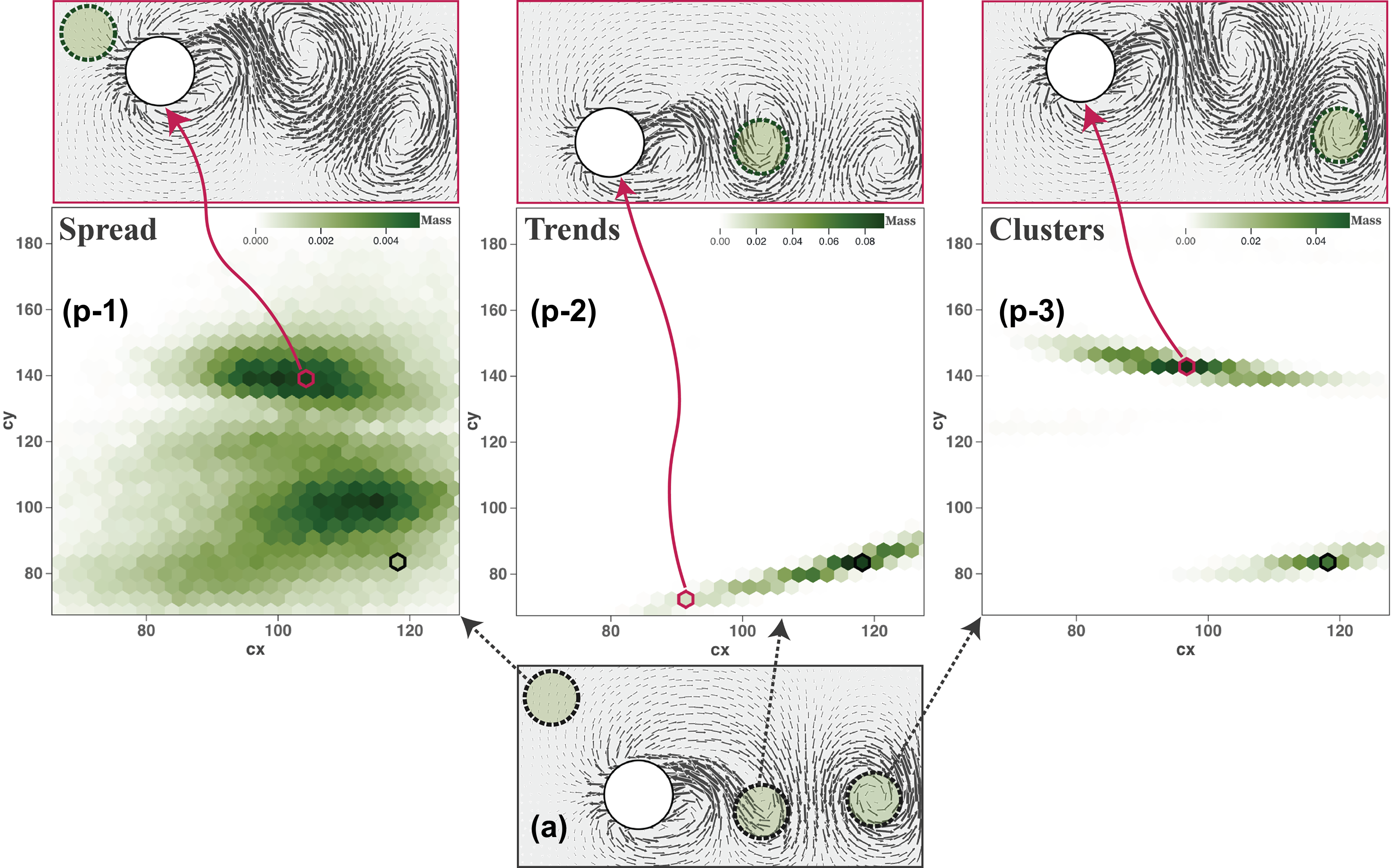}
    \caption{(a) Given features specified on a reference simulation, we show common patterns in our visualization design in preserving a feature: (p-1) spread of density conveys coverage of the parameter space, (p-2) trends indicate relationships between parameters, and (p-3) clusters convey distinct parameter configurations.}
    \label{fig:patterns}
\end{figure}

Last, we provide support for feature comparison.
Specifically, a user may specify a pair of features within a given field, leading to samples drawn from two distinct posterior distributions.
We then encode the two sets of samples via a bivariate color scale, where the addition of densities is encoded along luminance, dark and saturated colors encode a high density in one feature and low density in the other, and if the two features contain similar density, then a desaturated color will be assigned.

\subsection{Visual analysis}
\label{sec:analysis}

We provide some intuition for the types of visual patterns one should expect in our visualization, please see Fig.~\ref{fig:patterns}.
Specifically, we consider a fluid flow simulation in which the parameter space is 2-dimensional, consisting of the x and y coordinates of the center positioning of a circle obstacle, shown as a single heatmap.
Note to better emphasize features in the output, we subtract off the background horizontal motion in the flow.

\textbf{Spread of density} indicates the amount of coverage in the parameter space for which a feature is satisfied.
In Fig.~\ref{fig:patterns} (p-1), the selected feature corresponds to a region of laminar flow.
We thus find high density over much of the parameter space, with the exception of y coordinates assigned a large value, as this is where the circle obstacle would interfere with the specified neighborhood.
Note that density is not uniform.
This is a consequence of the prior, decreasing the density in regions of the parameter space in which it is more likely for the surrogate to incur error in its output.

\begin{figure}[!t]
    \centering
    \includegraphics[width=0.9\linewidth]{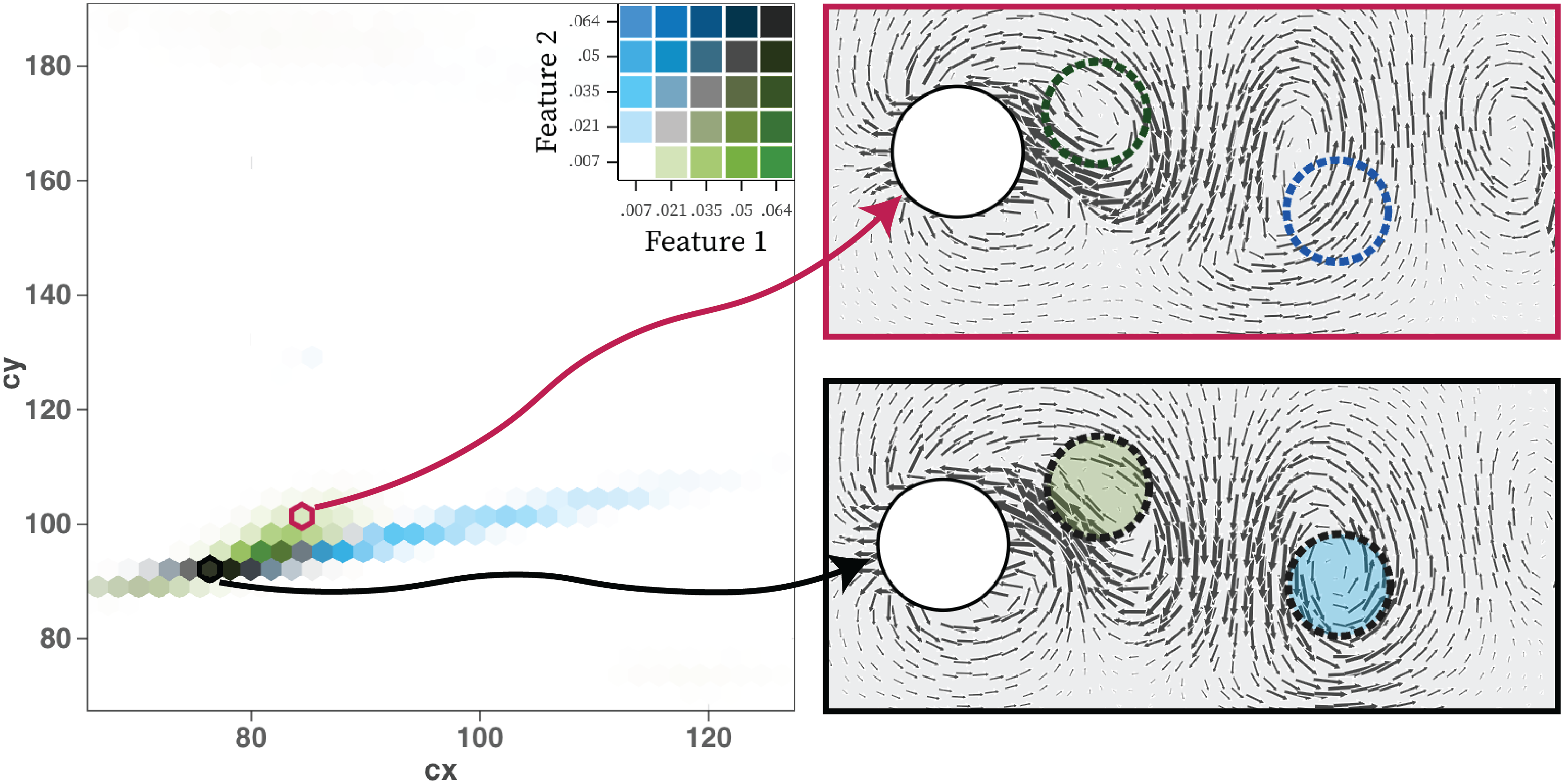}
    \caption{Upon specifying a pair of features, our visualization shows their corresponding distributions through a bivariate color scale, permitting a comparison of the distributions, namely their overlap, and where they differ (saturated colors).}
    \label{fig:comparison_patterns}
\end{figure}

\textbf{Trends} depict relationships between variables that are necessary to satisfy a feature.
Trends tend to manifest as 1-dimensional shapes within a subview of the matrix, whereby increasing the value of one variable requires an increase/decrease in the other.
In Fig.~\ref{fig:patterns} (p-2), preserving the specified vortex, an increase in the x-coordinate of the circle must be met with a (smaller) increase in its y-coordinate.

\textbf{Clusters} indicate distinct parameter configurations that lead to similar features in the output.
As shown in Fig.~\ref{fig:patterns} (p-3), clusters arise due to regions of high density being separated by regions of low density.
In this example, we specify a vortex-like feature towards the right side of the domain.
The heatmap shows that there are two main clusters, one for which the circle is positioned towards the upper-left of the domain, and the other for which the circle is positioned in the lower-right of the domain.

We show in Fig.~\ref{fig:comparison_patterns} representative visual patterns when comparing features.
The bivariate color scale maps Feature 1 intensity from light to dark green along the horizontal axis and Feature 2 intensity from light to dark blue along the vertical axis, with gray tones indicating overlap.
This color scale allows one to identify regions of the parameter space where the densities differ, namely where the colors are highly saturated, as well as their overlap, where colors are desaturated.
Typically, the reference simulation (black highlighted cell) serves as an ``anchor'', parameter configurations that are common to both features.
Moving away from this cell, we can identify differences in the features.
In this example, the green encoded feature corresponds to a vortex shed close to the circle obstacle, which occurs in a small portion of the parameter space.
In contrast, the blue color-encoded feature corresponds to a separate shed vortex, a feature that is common to a larger portion of the parameter space.

\section{Experimental results}
This section presents quantitative results on the effectiveness of our prior and HMC convergence, and qualitative results demonstrating our interface's advantages for parameter space exploration.

\begin{table}[!t]
\centering
\caption{Simulation specifications and dataset sizes. Resolution for \textsf{NACA airfoil} and \textsf{Rayleigh-Taylor instability} is adaptive in space and space-time, respectively, to sufficiently resolve features.}
{
\setlength{\tabcolsep}{4pt}
\footnotesize
\begin{tabular}{lcccc}
\toprule
Simulation & $d$ & Resolution & Train & Test \\
\midrule
\textsf{Circle} & 2 & $384 \times 256 \times 500$ & 100 & 900 \\
\textsf{NACA airfoil} & 3 & $1024 \times 256 \times 200^*$ & 300 & 700 \\
\textsf{Rayleigh-Taylor instability} & 4 & $128 \times 512 \times 248^*$ & 900 & 800 \\
\bottomrule
\end{tabular}
}
\label{dataset_table}
\end{table}

\paragraph{Datasets} We use three types of simulations for evaluating our method; please see Table~\ref{dataset_table} for additional details.
Specifically, the \textsf{Circle} simulation consists of a 2D parameter space representing the center coordinates of a circle obstacle embedded within a fluid flow, simulated using the WaterLily package~\cite{weymouth2023waterlily}.
The \textsf{NACA airfoil} simulation consists of a 3D parameter space that controls the shape of an airfoil embedded in a fluid, specifically its thickness, the camber (bend in the airfoil), and camber location (where the bend occurs).
The \textsf{Rayleigh-Taylor instability} simulation gives a 4D parameter space for a fluid mixing simulation, where a single parameter controls the density of one of the fluids (the denser one that transports down due to the force of gravity), and the remaining three parameters govern the shape of the initial interface separating the fluids.
Both \textsf{NACA airfoil} and \textsf{Rayleigh-Taylor instability} were run using the Gerris~\cite{popinet2004free}.
For both train \& test, parameter configurations are formed by sampling uniformly at random from the parameter space, where we train a neural field surrogate on the training parameters and use the test set for evaluation.

\subsection{Quantitative evalution}
\label{sec:quant_eval}

\subsubsection{Generalization of neural field surrogates}

\begin{figure}[!t]
    \centering
    \includegraphics[width=1.0\linewidth]{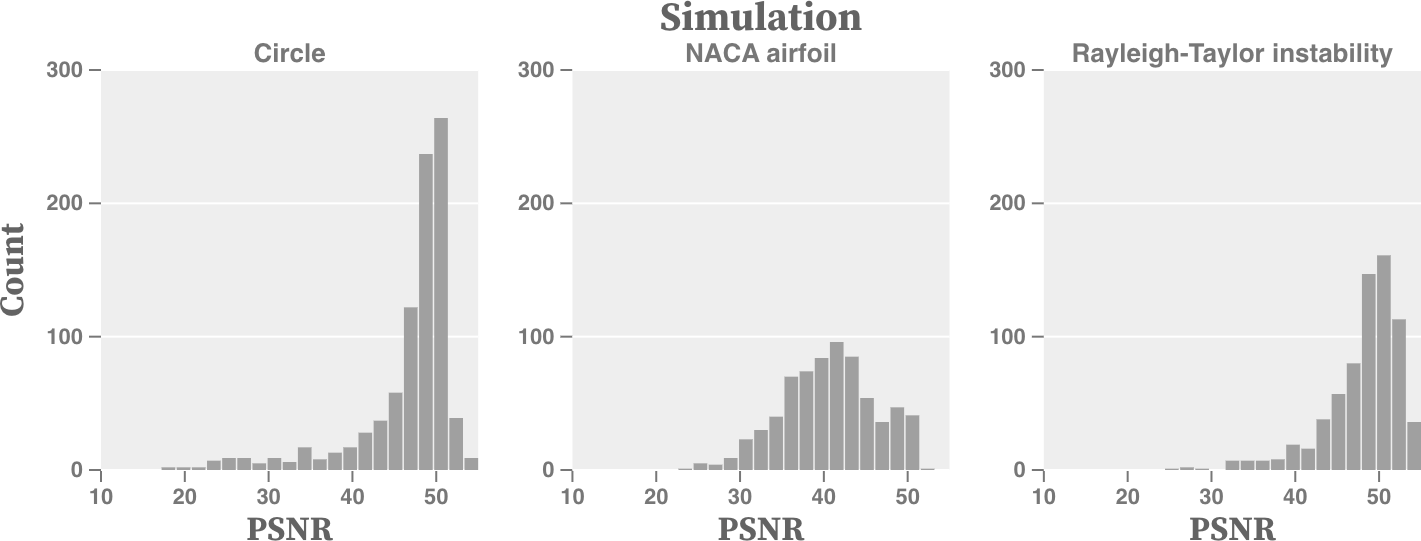}
    \caption{
    We evaluate the neural field surrogate's accuracy using a histogram of PSNR values across different simulations, focusing on parameters not seen during training. The results show good average performance, but left-tailed distributions indicate parameter configurations where models struggle to generalize.}
    \label{fig:acchist}
\end{figure}

Ideally, our trained neural fields exhibit good generalization.
Specifically, for parameter configurations not seen during training, model predictions match well with ground-truth simulation outputs.
Generalization quality will, at a minimum, be dependent on two factors: the number of simulations provided for training, and the complexity of the mapping (from input parameter to output field).
The latter factor is a property of the simulation, while the former can be controlled by the domain scientist/engineer.
In particular, as the number of simulation runs used for training increases, we should expect better generalization.
However, running many simulations can be quite time-consuming, especially for simulations that require detailed physics to be resolved.
Thus, when presented with a limited budget of simulation runs for training, the resulting neural field surrogate might not generalize well to arbitrary parameters, but rather just a subset of the parameter space.

\begin{figure}[!t]
    \centering
    \begin{subfigure}{1.0\linewidth}
        \includegraphics[width=1.0\linewidth]{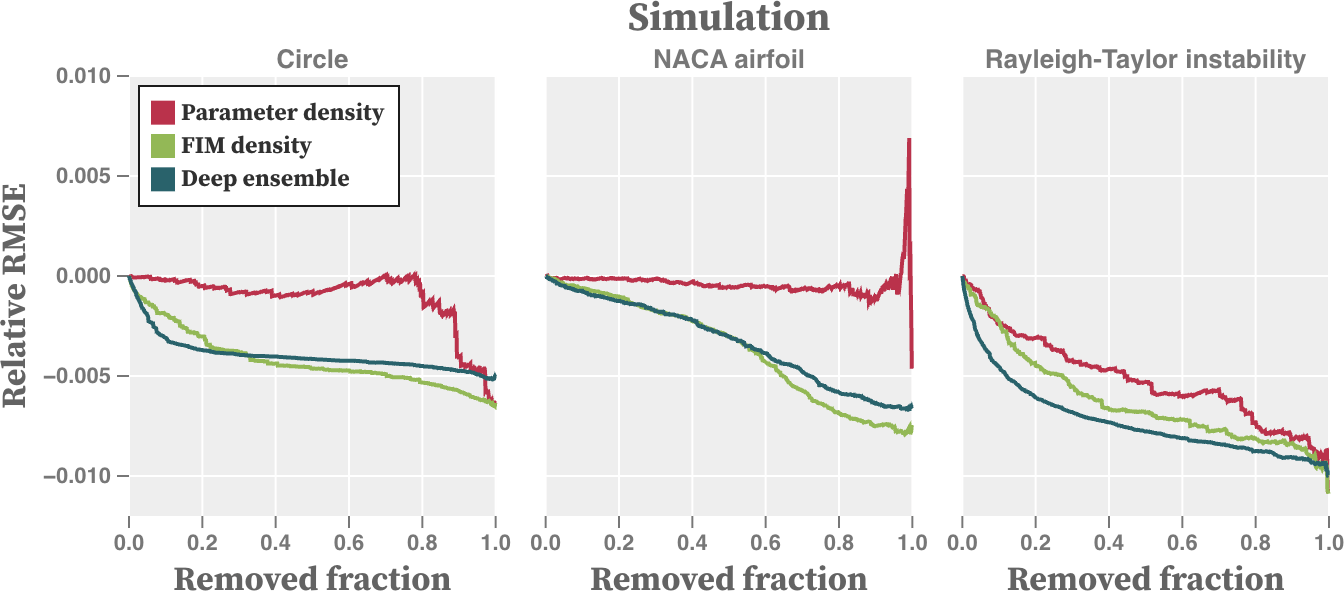}
        \caption{RMSE relative to evaluation on all parameters.}
        \label{fig:qualitative-good}
    \end{subfigure}
    \begin{subfigure}{1.0\linewidth}
        \includegraphics[width=1.0\linewidth]{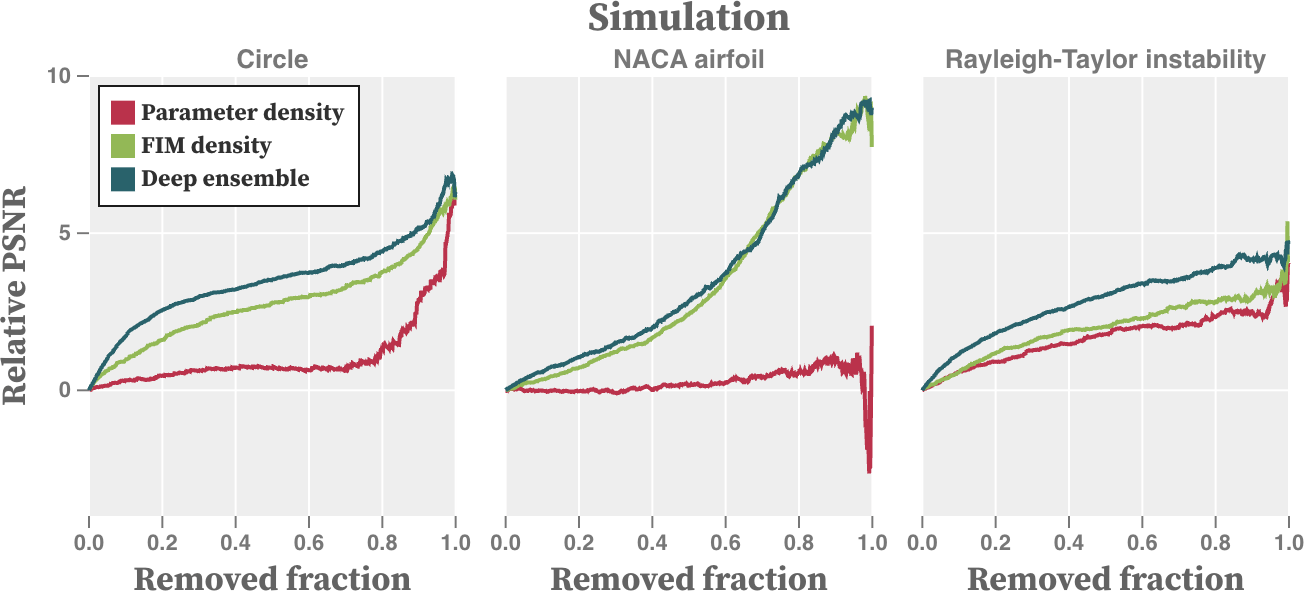}
        \caption{PSNR relative to evaluation on all parameters.}
        \label{fig:qualitative-bad}
    \end{subfigure}
    \caption{We compare our uncertainty measure with parameter-only KDE and deep ensembles. For each method we sort parameters by increasing likelihood, remove a specified fraction of parameters from the sorted list (varying on the x-axis), and average the RMSE/PSNR over all corresponding fields. We find that using our prior, the most likely parameters correspond to model predictions that are of higher accuracy, performing comparably to deep ensembles at a small fraction of the cost.}
    \label{fig:prior_eval}
\end{figure}

Fig.~\ref{fig:acchist} depicts histograms for peak signal-to-noise ratio (PSNR) across the simulations.
The data extents used in PSNR computation are simulation-specific.
We find that, across all simulations, model performance can vary, depending on the input parameters.
On one hand, the performance averaged over all withheld parameters is generally quite good, in the range of $40 - 45$ PSNR.
However, we observe a tail to the left in each of the histograms, indicating certain parameter configurations for which the surrogate poorly generalizes.
It is precisely these parameters that we should be assigned small prior density.
Specifically, though the surrogate might yield parameter configurations that match a user-supplied feature, these very parameters might give a poor approximation to the actual simulation outputs, and could thus mislead the end user.

\begin{figure}[!t]
    \centering
    \includegraphics[width=1.0\linewidth]{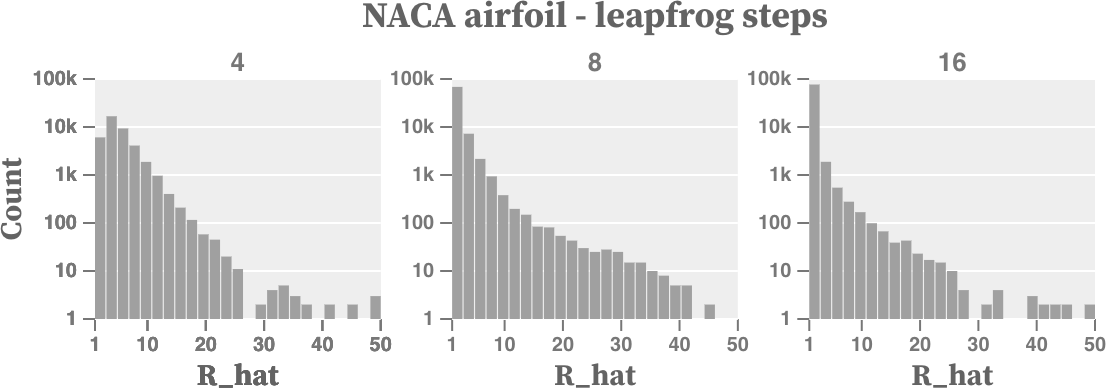}
    \caption{We show HMC convergence via the $\hat{R}$ statistic~\cite{gelman1995bayesian}, as a function of the number of leapfrog steps. We find 8 -- 16 steps leads to good mixing in chains.}
    \label{fig:rhat}
\end{figure}

\subsubsection{Effectiveness of prior}

We evaluate our proposed prior on parameters by adopting the evaluation protocol of sparsification plots~\cite{ilg2018uncertainty}.
Specifically, we sort parameters in increasing likelihood, remove a specified fraction (from 0 to 1) of parameters from this sorted list, quantitatively evaluate model predictions on the retrieved parameters via PSNR and root mean-squared error (RSME), and average the result.
Intuitively, the more parameters we remove, the higher the likelihood in what is retained, and thus we should expect the averaged performance to increase.
We find that our method gives this trend, whereas the parameter-only KDE tends to perform poorly.
We further compare to deep ensembles~\cite{lakshminarayanan2017simple}, where we independently train 10 SIRENs, and derive an uncertainty estimate via variance in the predictions, averaged over the field domain.
We find our method performs comparable to deep ensembles, but at a small fraction of the cost.
Specifically, our prior probability computation is approximately three orders of magnitude faster to compute.

\subsubsection{HMC convergence}

\begin{figure*}[!t]
    \centering
    \includegraphics[width=0.9\linewidth]{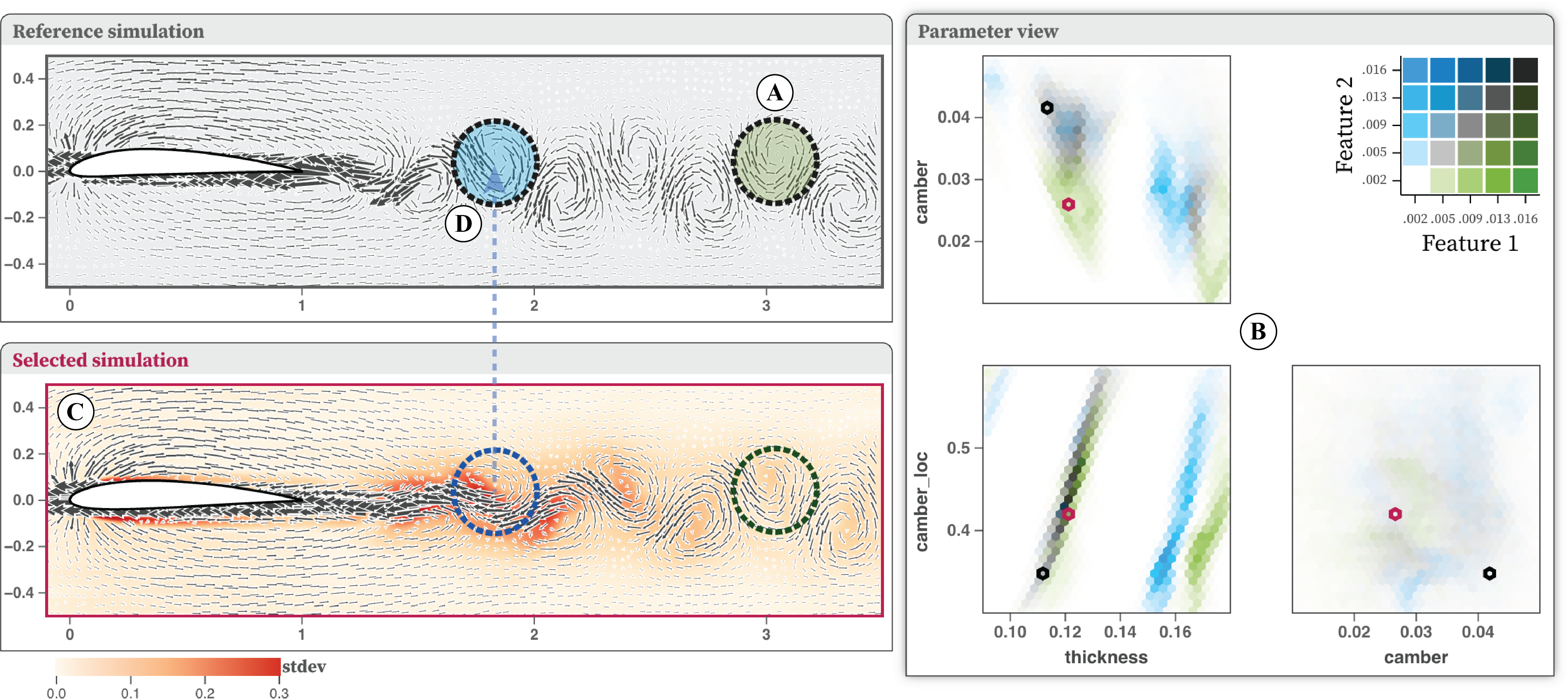}
    \caption{We illustrate our interface for visually analyzing the \textsf{NACA airfoil} simulation. A user first specifies a vortex as a feature (A), from which we run HMC to collect samples. The parameter view (B) depicts all configurations that give us the specified vortex. We further convey the variance in surrogate outputs over the set of collected samples (C), serving as guidance for selecting a second feature (D) which is expected to contrast with the first feature.}
    \label{fig:naca_airfoil_explore}
\end{figure*}

We next evaluate our parameter choices for HMC, namely the number of leapfrog steps used for integrating Hamilton's equations, please see the supplemental for additional convergence diagnostics.
The number of leapfrog steps dominates the computation time, and so this should be minimized while still ensuring mixing -- each chain has entered the intended stationary distribution.
We use the $\hat{R}$ statistic~\cite{gelman1995bayesian} as an indicator for mixing.
This computes the within-chain variance for a provided utility function, averaged over a set of independently-run chains, and measured relative to the variance over all chains.
A value close to 1 indicates good mixing.
For our utility functions we use the neural field evaluated over a set of spatial positions.
We compute $\hat{R}$ over a randomly-selected set of feature specifications, each dependent on a randomly chosen parameter configuration, and display the results as a histogram, varying over both the utility function and the specified feature.
Fig.~\ref{fig:rhat} shows results for the \textsf{NACA airfoil} simulation, please see supplemental material for results across all simulations, as well as a more exhaustive parameter comparison.
We find that when the number of leapfrog steps is set between 8 - 16, most $\hat{R}$ values are quite close to 1, indicating good mixing properties for chains.
Importantly, in this range of steps the latency in HMC is minimal -- advancing $50$ burn-in steps takes approximately 1 second, after which we progressively update heatmaps with collected samples along each chain.

\subsection{Parameter space exploration}
\label{sec:exploration}

We show the benefits of our method and visual interface in helping understand relationships between simulation outputs and parameter spaces.
We consider the \textsf{NACA airfoil} and \textsf{Rayleigh-Taylor instability} simulations in our experiments.
These simulations consist of nontrivial parameter spaces, e.g. of dimension 3 \& 4 respectively, where the link between features in the output and parameter variables is not apparent.
We further limit the results to feature comparison, in order to understand similarities/differences of features in terms of their respective posterior distributions.
Please see accompanying video for a more detailed illustration of our interface.

We first study the \textsf{NACA airfoil} simulation via a hypothetical user, please refer to Fig.~\ref{fig:naca_airfoil_explore}.
In this scenario, the user first inspects the simulation output for a reference parameter configuration whose values are shown via the black cells in the Parameter view.
They specify a spatial neighborhood via dragging the green circle within the field (A), corresponding to a vortex shed from the airfoil.
Their specification prompts a call to our HMC sampler, where upon completion of a fixed number of burn-in steps, the binned heatmaps in the Parameter view progressively update to show the steps made along a batch of chains (B).
At the conclusion of sampling, the Selected simulation view is updated to show the variance of the surrogate outputs over posterior samples, drawn as a color map (C) -- note the high variance regions closer to the tip of the airfoil.
To contrast features, the user specifies another spatial neighborhood corresponding to a vortex for which the variance is high, namely the blue circle in the Reference simulation (D).
The HMC sampler is then run for this feature, and the Parameter view is updated to show both distributions under our bivariate color scale.
To see where these distributions differ, the user selects a simulation within the Parameter view corresponding to the highlighted red cells.
This gives a parameter configuration that has high posterior density for the first (green) feature, but low density for the second (blue) feature.

For this scenario we make a number of observations.
First, inspecting the Parameter view, we find a large density spread for the marginal distribution of camber \& camber location, indicating little relationship between these variables in reproducing the specified features.
Rather, we see clear relationships between thickness, and both camber \& camber location due to the peaked density patterns.
Specifically we see various trends between thickness and camber location -- in general, an increase in thickness requires a corresponding increase in the camber location to give the features.
We further find clusters in these plots that suggest differences in the posterior densities, e.g. regions of saturated blue and green colors.
The selected parameter configuration, e.g. selected red cells in the Parameter view, indicates a configuration that is most distinguished by the camber.
Here we see that decreasing the camber, e.g. a reduction in the bend of the airfoil, leads to a suppression of the vortex corresponding to the blue-colored feature, while preserving the specified vortex that has occurred later in the simulation.
Essentially, this behavior shows that curved airfoils produce vortex sheds in their immediate vicinity, whereas straighter (less camber) airfoils have their vortex shed form more slowly.

\begin{figure*}[!t]
    \centering
    \includegraphics[width=0.8\linewidth]{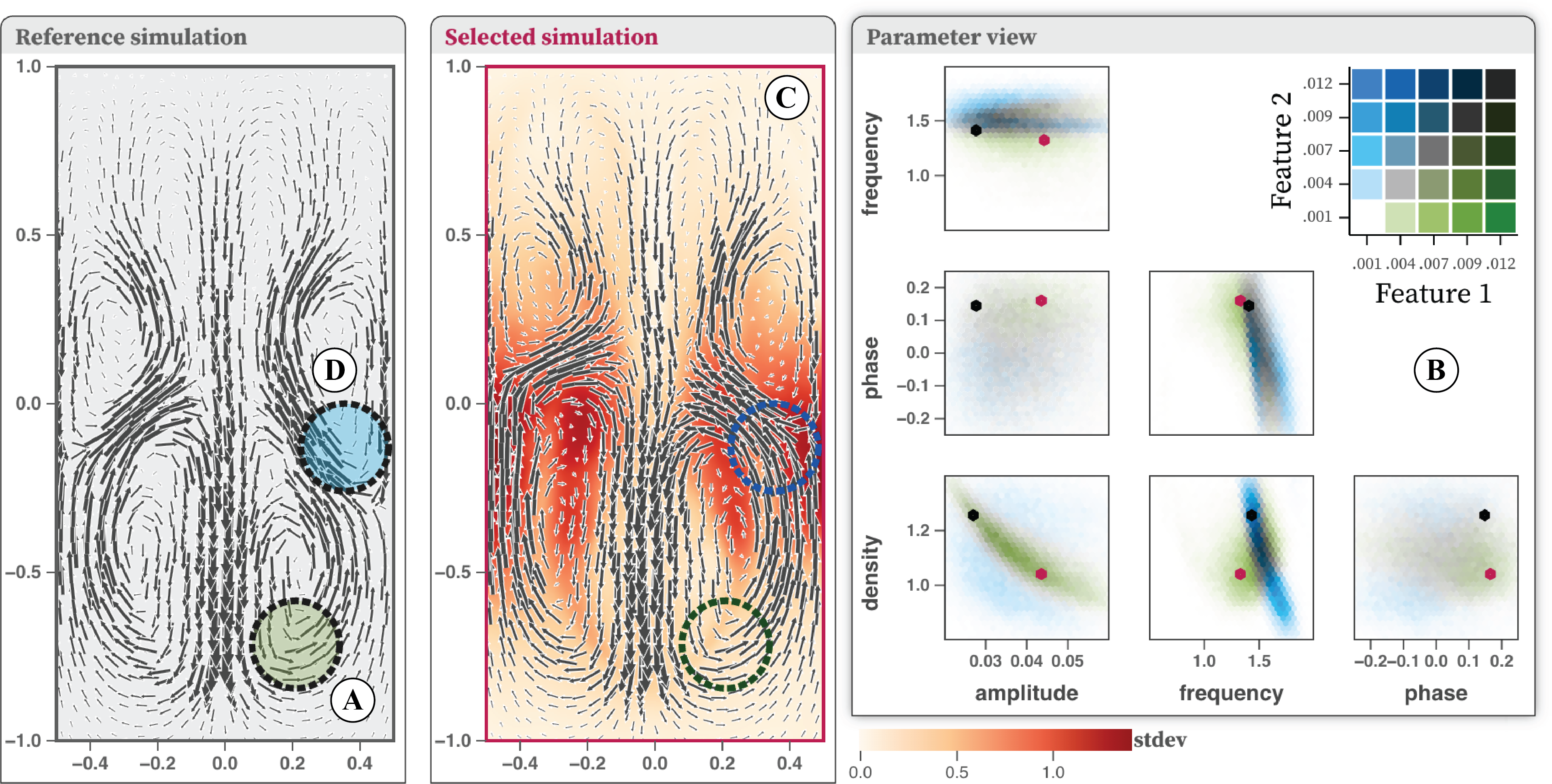}
    \caption{We analyze the parameter space of a \textsf{Rayleigh-Taylor instability} simulation. Here we compare features corresponding to rotational flow, shown via green (A) and blue (D) circles. Comparing their corresponding distributions (B) highlights portions of the parameter space that they have in common, as well as regions that distinguish the features, further supported in the output variance (C).}
    \label{fig:rt_explore}
\end{figure*}

We next study the \textsf{Rayleigh-Taylor instability} simulation, see Fig.~\ref{fig:rt_explore}.
Here the user inspects the Reference simulation, corresponding to a chosen parameter configuration, to identify a feature of interest.
The specified green circle (A) represents the counter-clockwise rotation of flow.
Running HMC, we obtain a collection of samples displayed in the binned heatmaps (B) as well as color-mapped variance (C).
The user identifies another region of rotational flow (blue circle), but one that contains high variance (D), from which samples of the posterior distribution are drawn and overlaid with the other feature's distribution.
The user then selects a configuration from the Parameter view (red cells) corresponding to parameter values that give the bottom rotational feature (green), but exclude the middle rotational feature (blue).

The visual patterns in this scenario allow us to make several observations.
First, the Parameter view shows that the marginal distribution for amplitude and density leads to a nonlinear trend for the green-colored feature, whereas for the blue-colored feature, the marginal indicates little relationship between these variables.
For the green feature, we observe a compromise between amplitude and fluid density.
An increase in the density of the top fluid facilitates mixing, but we achieve a similar effect via an increase in amplitude.
A large amplitude gives a more pronounced ``pinch'' in the initial interface between fluids, thus facilitating transport in the top fluid.
Thus in order to preserve the specified counter-clockwise flow, an increase in amplitude must be met with a decrease in fluid density, leading to the depth (y coordinate) at which mixing occurs to be largely unchanged.
We additionally find that the parameter variables which best contrast the distributions are frequency and density.
A decrease in frequency from the reference simulation (black cells in Parameter view) leads to configurations that place us out-of-distribution in both feature posteriors.
To place us back in-distribution for the green-colored feature, we must also decrease the density -- yet in doing so, the blue-colored feature is no longer preserved.
This reveals a stronger dependency on the shape of the interface for the blue feature -- a change in flow circulating towards the top -- compared to the green.
This is expected, since the location of the blue circle is placed closer to the interface, whereas the feature of counter-clockwise rotation occurs towards the bottom, and thus preserved across a range of frequency \& density values.

\section{Discussion}
We have introduced a method for analyzing simulation surrogates, combining Bayesian inference with neural fields.
Through visualizing a distribution of parameter configurations, we show it is possible to gain useful insights into simulation parameter spaces.
This supports exploratory analyses such as identifying trends, correlations, and patterns among simulation inputs and their outputs.

We acknowledge several limitations with our method.
First, our current study considers up to four parameter variables. In very high-dimensional spaces, the samples needed for accurate posterior approximation may grow prohibitively, and our matrix-based view becomes impractical. While the method remains applicable in principle, effective use would require strategies such as view informativeness criteria (e.g., Scagnostics~\cite{wilkinson2005graph} adapted to marginal density plots) or alternative designs using dimensionality reduction.
Secondly, a central concern in any surrogate-driven workflow is the reliability of the surrogate itself.
When the input‑to‑output relationship is highly intricate, a SIREN surrogate might struggle to generalize over the full parameter space.
Crucially, the surrogate is isolated from our sampler and visualization modules,
so many state-of-the-art surrogates~\cite{muller2022instant,chen2025explorable} can be substituted with minimal integration cost.
Last, throughout the paper, we treat a feature as a 2-D spatial patch at a single time step, deliberately setting aside other potentially more useful features, such as gradient-based quantities, integral curves, and other spatio-temporal aggregates.
Further usability challenges arise when moving to volumetric fields as simulation outputs.
To extend our method for richer feature comparisons, each feature should be formulated as a differentiable quantity of the neural field surrogate, accompanied by interaction designs that enable specification of these features.

For future work, we aim to generalize inference to both parameter inputs and space-time locations, allowing one to retrieve features over the full domain of the simulation.
We also aim to consider different ways of expressing features, e.g. one might be interested in suppressing a given feature, rather than preserving it.
We intend to investigate the broader framing of feature specifications in surrogate outputs.
Last, our approach emphasized parameters that, both, satisfy a given feature, and are likely to give good approximations to simulations.
It would be worth decoupling these quantities, so that one can investigate model uncertainty separately from feature-based querying.

\acknowledgments{
This research was supported by the U.S. Department of Energy under grants DE-SC0023320 (Vanderbilt University) and DE-SC-0023319 (The University of Arizona).
}

\bibliographystyle{abbrv-doi}

\bibliography{vis,matt-refs}

\clearpage

\appendix

\section{Implementation details}

\subsection{Neural field surrogate}

Our neural field surrogate takes the form of a SIREN~\cite{sitzmann2020implicit}, namely a simple multi-layer perceptron (MLP) whose inputs consist of both spatiotemporal coordinates from the field domain, as well as the values of all parameter variables.
Across all simulations an MLP consists of 4 hidden layers, each of width $256$.
We use Adam~\cite{kingma2014adam} for optimization with a minibatch size of $100,000$, and take $60,000$ optimization steps.
We set a base learning rate of $0.0002$, linearly annealed over the course of optimization.
For our datasets, since all simulation runs cannot be stored in memory, during optimization we collect a random sampling of coordinates in the domain for a random selection of simulations, enough to fill CPU memory.
This process is repeated once we have made a pass over all of the stored data, thus ensuring that both spatiotemporal coordinates and parameter configurations are sampled uniformly at random.
The training time for our models, in large part independent of simulation, amounts to approximately 9 minutes.
The elapsed time to generate simulation outputs used in training for \textsf{NACA airfoil} and \textsf{Rayleigh-Taylor instability} is, respectively, 36 hours and 26 minutes, and 9 hours and 29 minutes.
Synthesizing a vector field at a given time step and parameter configuration can be done in an interactive manner, using our SIREN model with the aforementioned specification, as the accompanying video demonstrates.

A central consideration to a SIREN is the scale applied to the inputs~\cite{tancik2020fourier}, a necessity in allowing the network to reproduce high-frequency details in the output.
Although it is convention to apply a scale of $30$~\cite{sitzmann2020implicit}, we find that applying this to both spatiotemporal coordinates $\mathbf{x} \in \mathcal{X}$ and parameter values $\mathbf{z} \in \mathcal{Z}$ leads to overfitting.
Through experimentation, we find this scale applied to $\mathbf{x}$ is warranted, but rather, the source of overfitting is this scale applied to $\mathbf{z}$.
Hence we apply a scaling specific to parameter values $\mathbf{z}$, in practice we set this to $10$ in all of our models.
We find this to be large enough to capture details, while preventing overfitting to the training data.

\subsection{Density estimation}

For our density estimate, an estimate of the Fisher information matrix (FIM) requires sampling spatiotemporal coordinates over the domain.
As a preprocess, we compute the FIM corresponding to each training parameter configuration, where we take $2^{20}$ samples over the field's domain for estimation.
At inference time, the density estimate thus only requires the evaluation of a Mahalanobis distance in the parameter domain between all training parameters, an inexpensive operation.
Since HMC requires the gradient of log (unnormalized) densities, for numerical stability we apply the logsumexp trick~\cite{blanchard2021accurately}.

\begin{figure}[!t]
    \centering
    \includegraphics[width=1.0\linewidth]{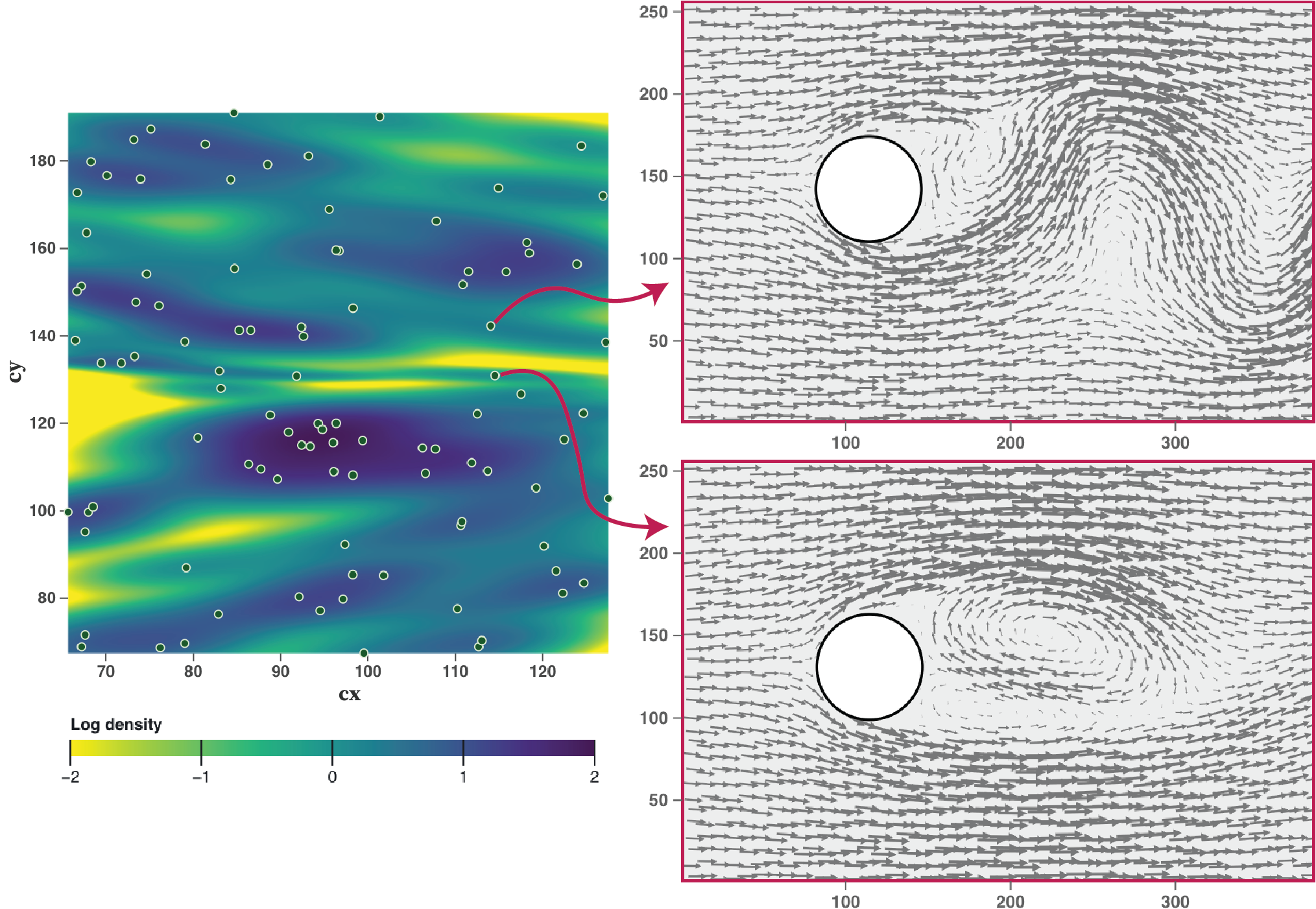}
    \caption{We illustrate our proposed density estimator, evaluated over the parameter space of the \textsf{Circle} simulation, shown as a heatmap (left). Circles in the plot correspond to training data parameter configurations. For two configurations separated by low density, we find the field rapidly changing (left) when moving from one configuration to the other.}
    \label{fig:fim}
\end{figure}

\subsection{HMC}

We employ HMC~\cite{gelman1995bayesian, neal2011mcmc} to sample from the posterior, which is a Markov Chain Monte Carlo (MCMC) method designed to sample efficiently from complex, high-dimensional probability distributions.
HMC simulates Hamiltonian dynamics to propose new states by integrating the system’s equations of motion using the leapfrog method.
This enables exploration of the parameter space along trajectories that follow the geometry of the target distribution, reducing the random walk behavior common in traditional MCMC methods.

\begin{figure*}[!t]
    \centering
    \begin{subfigure}{1.0\linewidth}
        \includegraphics[width=1.0\linewidth]{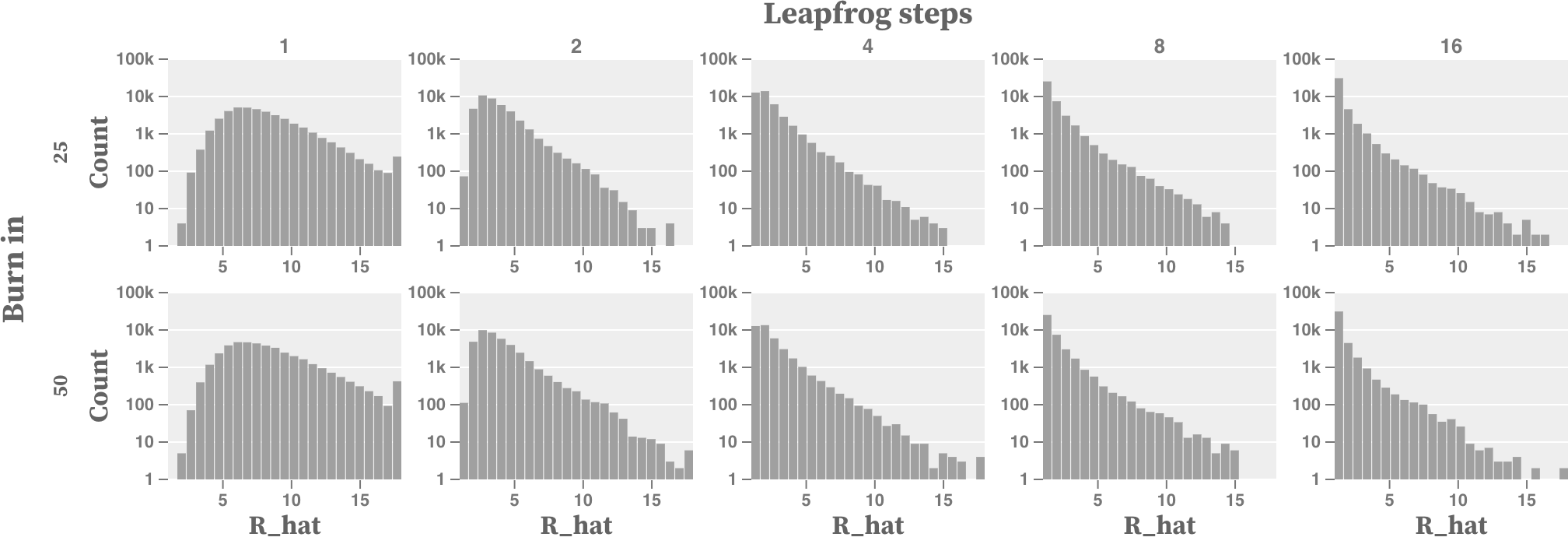}
        \caption{\textsf{Circle}}
    \end{subfigure}
    \begin{subfigure}{1.0\linewidth}
        \includegraphics[width=1.0\linewidth]{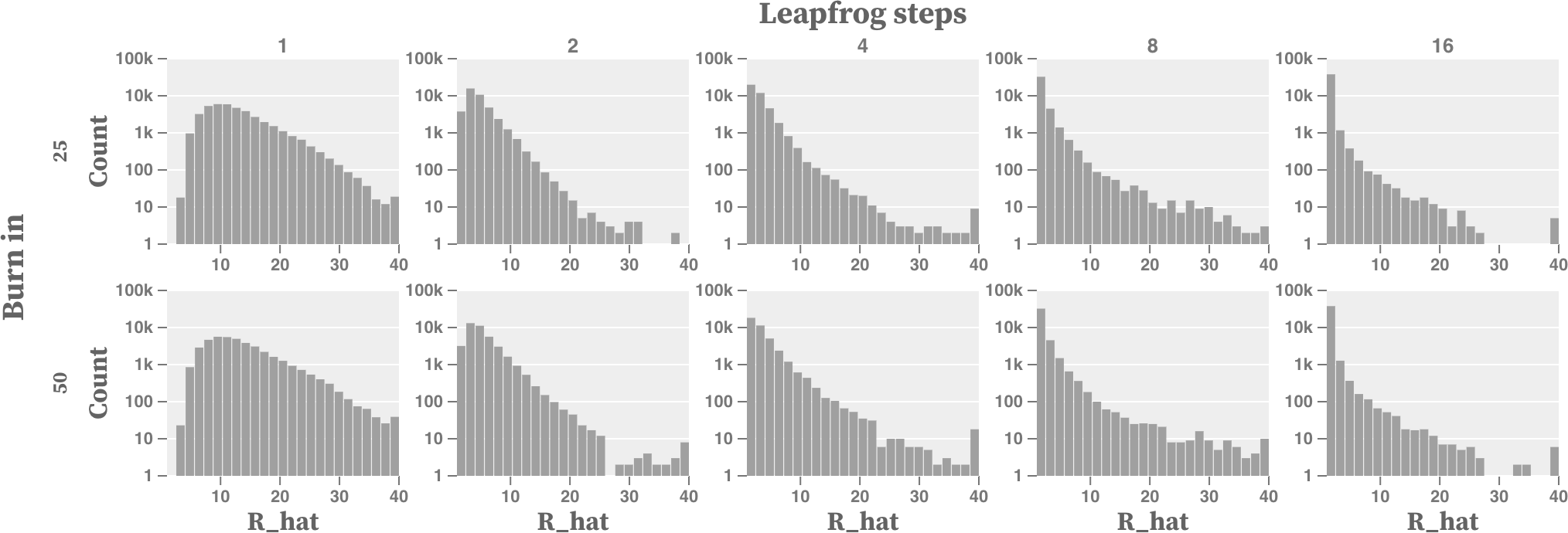}
        \caption{\textsf{NACA airfoil}}
    \end{subfigure}
    \begin{subfigure}{1.0\linewidth}
        \includegraphics[width=1.0\linewidth]{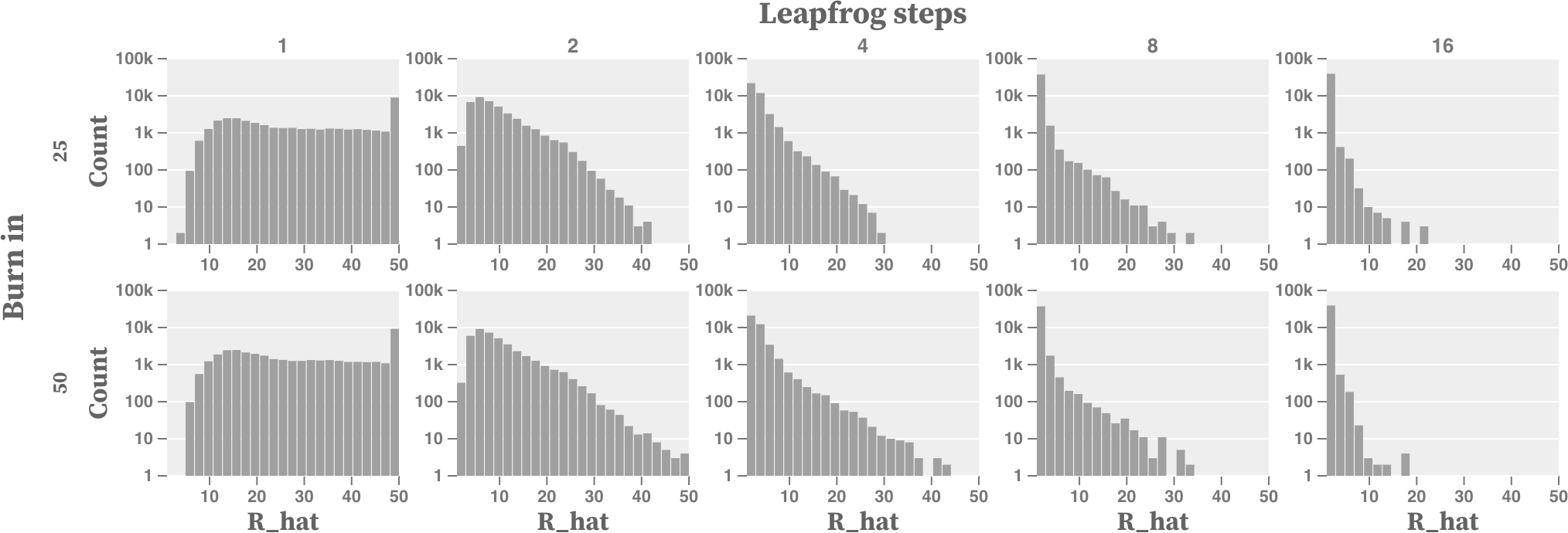}
        \caption{\textsf{Rayleigh-Taylor instability}}
    \end{subfigure}
    \caption{$\hat{R}$ statistic, presented as a histogram that varies over different feature specifications and surrogate outputs. We study the effect of (1) the number of burn-in steps for HMC, as well as (2) number of leapfrog steps. We find small differences in burn-in, while showing that at 8 - 16 leapfrog steps we obtain small $\hat{R}$ values, indicating that most chains exhibit mixing.}
    \label{fig:rhat}
\end{figure*}

The simulation uses two primary parameters: (1) Step size — controls the integration accuracy; (2) Number of leapfrog steps — determines how far the trajectory travels before a Metropolis acceptance/rejection step.
By following these simulated trajectories, HMC can generate proposals that are far from the current state but still have a high acceptance probability, leading to faster mixing and more efficient sampling.

We run HMC, in parallel, over a batch of chains -- we set the batch size to $1,000$.
HMC is run in two phases.
First, we run for a fixed number of burn-in steps, set to $50$ to ensure sufficient time for mixing.
Afterwards, we progressively update the heatmap with the results of HMC.
Specifically, for every $5$ steps of HMC, we collect the set of samples and update the binned heatmaps.
This is done for $20$ iterations, thus at conclusion giving $100$ steps of HMC.
In practice, samples might exit the parameter domain $\mathcal{Z}$; although our density estimate penalizes this, it is nevertheless possible.
We exclude parameter configurations that leave the parameter domain.
In total, HMC gives at most 100K samples that we display in our visual interface.

As part of HMC, we require evaluating the (log) of the feature likelihood for any $\mathbf{z} \in \mathcal{Z}$ with respect to user-specified reference $\mathbf{\hat{z}} \in \mathcal{Z}$.
This requires a sampling over a specified spatial region $X \in \mathcal{X}$:
\begin{equation}
d_X(\mathbf{z}, \mathbf{\hat{z}}) = \frac{1}{K} \sum_{k=1}^K \lVert f(\mathbf{x}_k, \mathbf{z}) - f(\mathbf{x}_k, \mathbf{\hat{z}}) \rVert,
\label{eq:nll}
\end{equation}
with each $\mathbf{x}_k \in X$.
A denser sampling, e.g. large $K$, leads to a more accurate approximation of this neighborhood-based feature, yet in turn computation of gradients comes at a high expense.
To this end, at each leapfrog step of HMC, we sample a fresh set of $K$ locations uniformly at random in the domain $\mathbf{x}_k \in X$, where in practice we set $K$ to 30.
This introduces stochasticity in the feature likelihood.
However, it is an unbiased estimate of the feature likelihood integrated over all of $X$, and ensures that the likelihood is not fixed to a predefined set of positions.

\begin{figure}[!t]
    \centering
    \begin{subfigure}{0.98\linewidth}
        \includegraphics[width=1.0\linewidth]{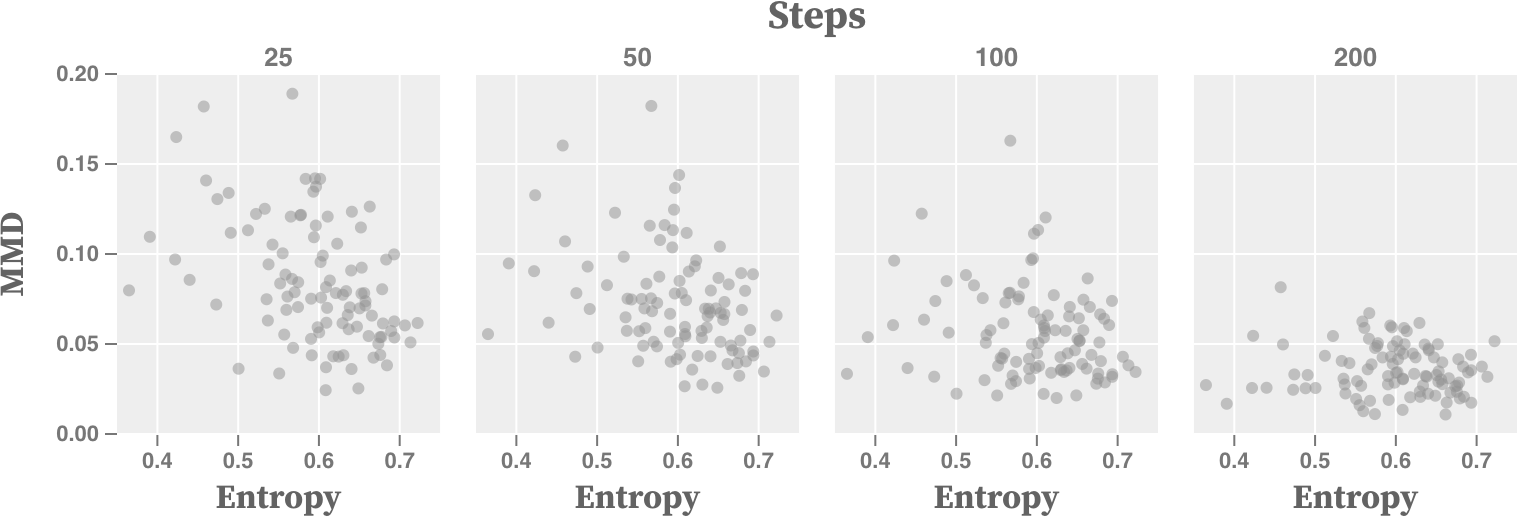}
        \caption{\textsf{Circle}}
        \label{subfig:mmd_circle}
    \end{subfigure}
    \begin{subfigure}{0.98\linewidth}
        \includegraphics[width=1.0\linewidth]{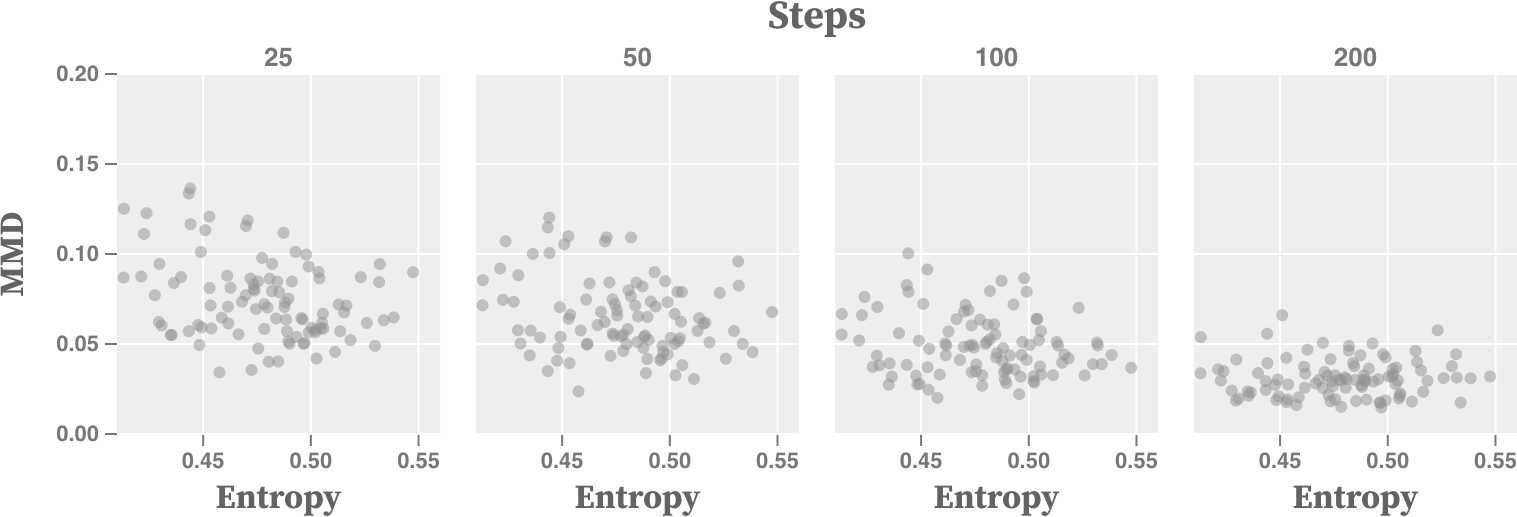}
        \caption{\textsf{NACA airfoil}}
        \label{subfig:mmd_naca_airfoil}
    \end{subfigure}
    \begin{subfigure}{0.98\linewidth}
        \includegraphics[width=1.0\linewidth]{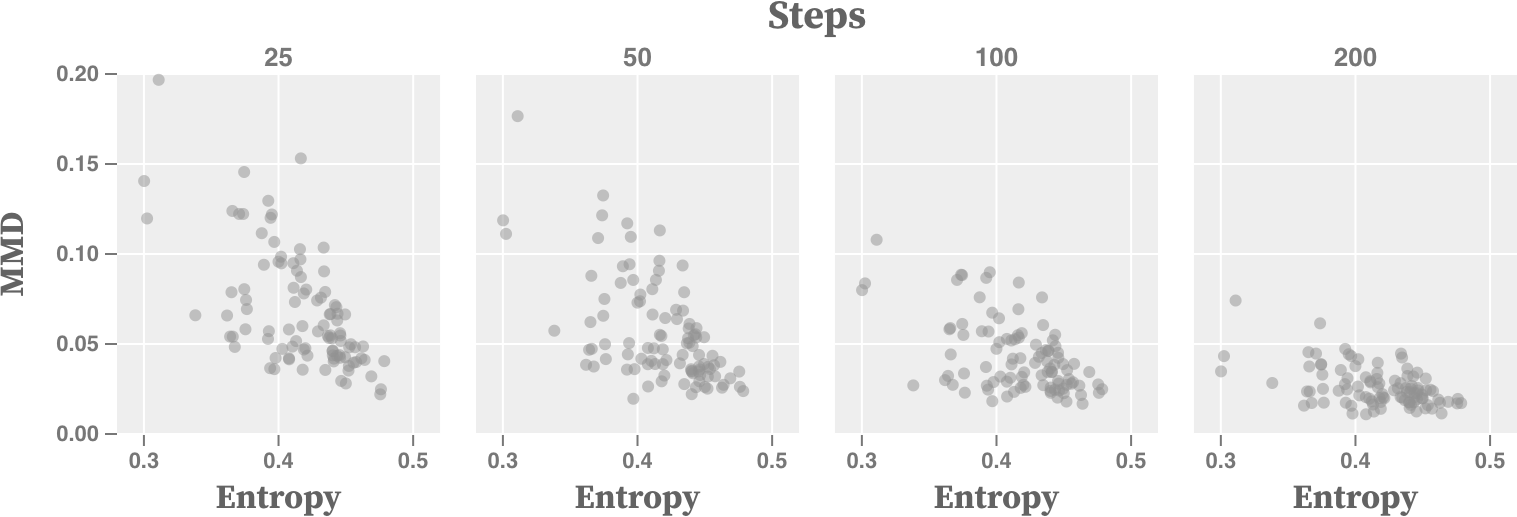}
        \caption{\textsf{Rayleigh-Taylor instability}}
        \label{subfig:mmd_rt}
    \end{subfigure}
    \caption{We show convergence of HMC by varying the number of steps taken for a set of chains, comparing against a reference HMC result of 400 total steps, measured by MMD. Each circle in the scatterplot corresponds to a feature specification made in a random position, for a random parameter configuration, where we plot the entropy of the reference HMC against MMD. We find good convergence between 100 - 200 steps, giving low latency in the collection of samples.}
    \label{fig:mmd}
\end{figure}

\subsection{Feature likelihood}

We recall that we adopt an energy-based model for our feature likelihood:
\begin{equation}
    p(X | \mathbf{z}) \propto \exp\left(-\frac{d_X(\mathbf{z}, \mathbf{\hat{z}})}{C}\right).
    \label{eq:spec}
\end{equation}
A crucial parameter is the value of $C$.
In practice, we have fixed $C = \frac{1}{15}$ across all simulations.
This is motivated by the simulations we have used, where the average vector norms between simulations is around 1, largely a consequence of the initial conditions being unit-norm vectors.
Thus, considering the NLL (c.f. Eq.~\ref{eq:nll}), with this setting of $C$ if the average distance in vectors is $0.2$ then Eq.~\ref{eq:spec} will report a value of approximately 0.05, and exponentially decay as the NLL increases.
We find at this averaged distance that feature matches tend to make sense as reasonable approximations.

\section{Density estimation illustration}

To illustrate that our proposed density estimator captures a notion of field proximity, we take the \textsf{Circle} simulation and evaluate the density on a dense regular grid that covers the parameter space, namely, the center position of the circle obstacle.
As shown in Fig.~\ref{fig:fim}-left via log density, the streaks of yellow (low density, or high uncertainty) tend to horizontally span certain y coordinate values.
This anisotropy reflects the simulation's larger sensitivity to changes in the y coordinate of the circle, in comparison to its x coordinate.
The circles in the plot correspond to parameter configurations used for training the model, where we highlight a pair of configurations divided by a region of low density.
Inspecting the corresponding vector fields (right), we find that the circle whose y coordinate is near the vertical center of the domain gives a simulation (bottom) in which the Kármán vortex street is yet to form at this particular time step.
In contrast, moving the y coordinate slightly up, we see that the vortex street has formed.
This is a consequence of the vertical forces of the flow more easily canceling out when the circle's y coordinate is closer to the center, thus requiring more effort to produce the vortex street.
Our density estimator is thus able to capture a notion of field proximity, namely, rapid changes in the field are more likely to lead to lower density.

\begin{figure}[!t]
    \centering
    \includegraphics[width=1.0\linewidth]{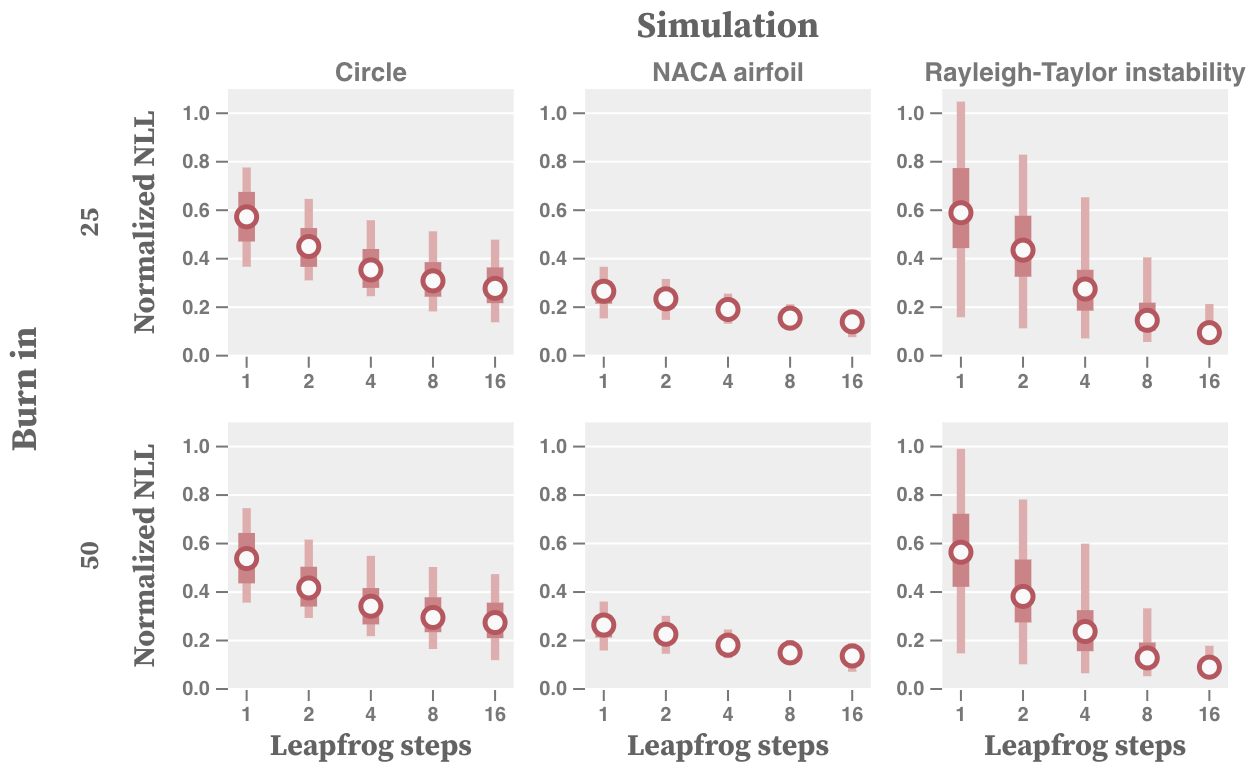}
    \caption{We show the distribution of normalized negative log likelihood (c.f. Eq.~\ref{eq:nll}) for samples collected from HMC, varying over number of leapfrog steps and burn-in iterations. We find that 8 - 16 leapfrog steps gives good NLL values, indicative of approximate feature matches.}
    \label{fig:energy}
\end{figure}

\begin{figure}[!t]
    \centering
    \includegraphics[width=1.0\linewidth]{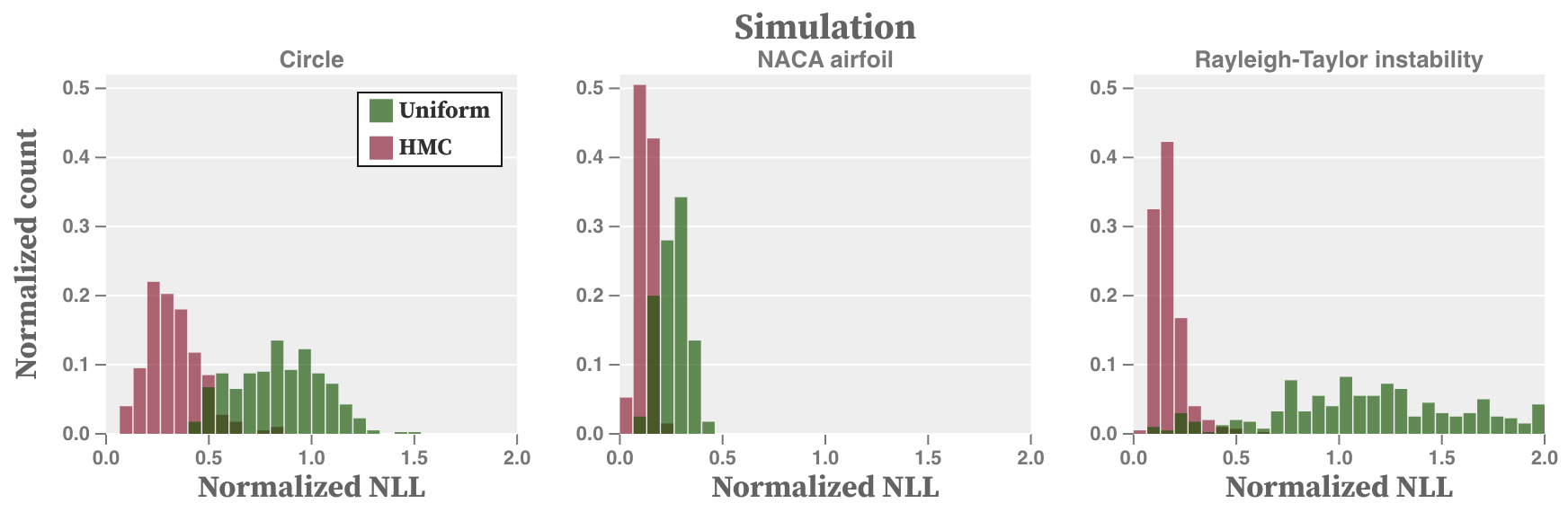}
    \caption{We compare the quality of the feature matches retrieved through HMC, at 16 leapfrog steps, against parameters sampled uniformly at random over the parameter domain.}
    \label{fig:energy_compare}
\end{figure}

\section{HMC diagnostics}

We include additional results related to choices made in HMC.
First, expanding on the $\hat{R}$ statistic results in the paper, we provide a more comprehensive set of experiments that vary over a wider range of leapfrog steps, as well as varying the burn-in steps -- from 25 to 50.
Please see Fig.~\ref{fig:rhat} for results.
We find that in going from 4 to 8 leapfrog steps we see a rather large change in histograms, indicating that at 8 leapfrog steps we are able to obtain good mixing.
Moreover, we find that at 25 burn-in steps the results are comparable to that of 50, yet to just ensure convergence we stick with 50 burn-in steps in our implementation.

We additionally study our choice of taking 100 steps of HMC, for a batch of 1,000 chains.
In particular, we are interested in seeing whether this gives good coverage in the parameter space.
Specifically, we are not just concerned with drawing samples from the posterior, but rather ensuring we have drawn enough samples such that the density of samples serves a good approximation to the actual posterior density.
To study this matter, we vary the number of steps taken in HMC, comparing against a reference HMC solution of $400$ steps.
We aim to see at what number of steps does HMC converge to the reference.
We measure the similarity between two sets of samples using maximum mean discrepancy (MMD)~\cite{gretton2012kernel}, which can be understood as a generalization of a test statistic to samples drawn from high-dimensional probability distributions.
Please see Fig.~\ref{fig:mmd} for results, where we plot the estimated entropy of the reference samples against MMD.
We find that in 100 - 200 steps we obtain good convergence.
We further find that higher-entropy distributions require fewer number of steps for convergence.
Low-entropy distributions tend to be concentrated in a rather small region of the parameter space; as such, they require finer precision in the placement of samples to obtain a good approximation.

We next show the quality of the feature matches found from our HMC sampler.
Specifically, in Fig.~\ref{fig:energy} we show the negative log likelihood (NLL), specifically Eq.~\ref{eq:nll}, normalized via dividing by the average vector norm in the respective simulations.
Consistent with the previous results, we find 8 - 16 leapfrog steps gives small NLL values; visual inspection confirms that such values lead to good approximate feature matches.
We further find a small decrease in NLL when increasing the number of burn-in iterations, though it is relatively minor.
Last, Fig.~\ref{fig:energy_compare} shows that the parameter configurations gathered through HMC gives improved feature matches over naive uniform sampling of the parameter space.

\end{document}